\documentclass[11pt,a4paper]{article}
\usepackage{acl2023}

\usepackage{times}
\usepackage{url}
\usepackage{latexsym}
\usepackage{microtype}
\usepackage{amsmath}
\usepackage{amsfonts}
\usepackage{graphicx}
\usepackage{adjustbox}
\usepackage{booktabs}
\usepackage{subcaption}
\usepackage{multirow}
\usepackage[safe]{tipa}

\usepackage{inconsolata}

\usepackage{cleveref}
\crefname{section}{\S}{\S\S}
\Crefname{section}{\S}{\S\S}
\crefname{table}{Table}{}
\crefname{figure}{Figure}{}
\crefname{algorithm}{Algorithm}{}
\crefname{equation}{Equation}{}
\crefname{appendix}{Appendix}{}
\crefformat{section}{\S#2#1#3}  

\newcommand{\word}[1]{{\em #1}}
\newcommand{\att}[2]{\textrm{\textsf{#1}=\textsc{#2}}}

\newcommand{\PL}{{\setlength{\fboxsep}{0pt}\fbox{\includegraphics[height=0.30cm,width=0.45cm]{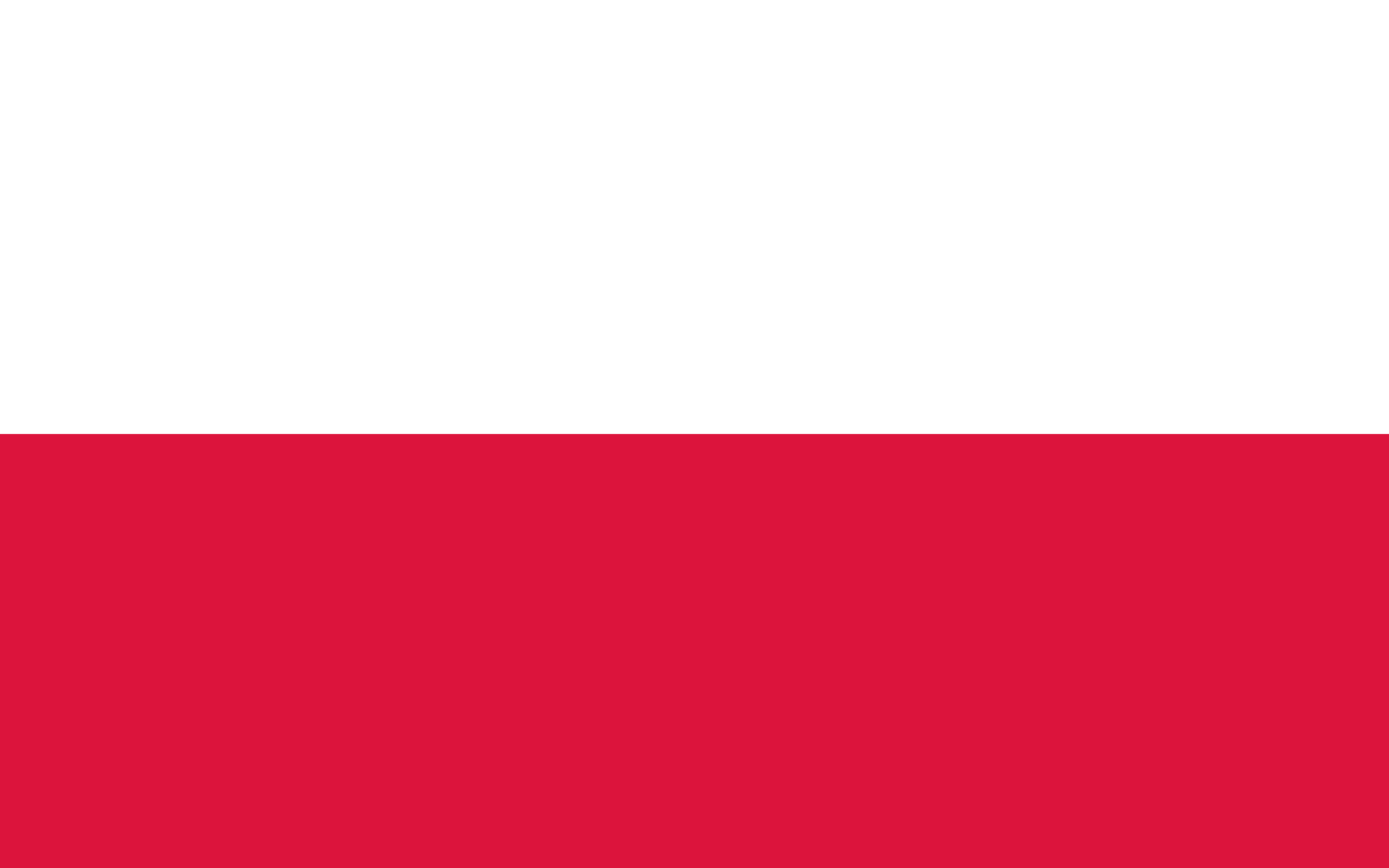}}}}
\newcommand{\RU}{{\setlength{\fboxsep}{0pt}\fbox{\includegraphics[height=0.30cm,width=0.45cm]{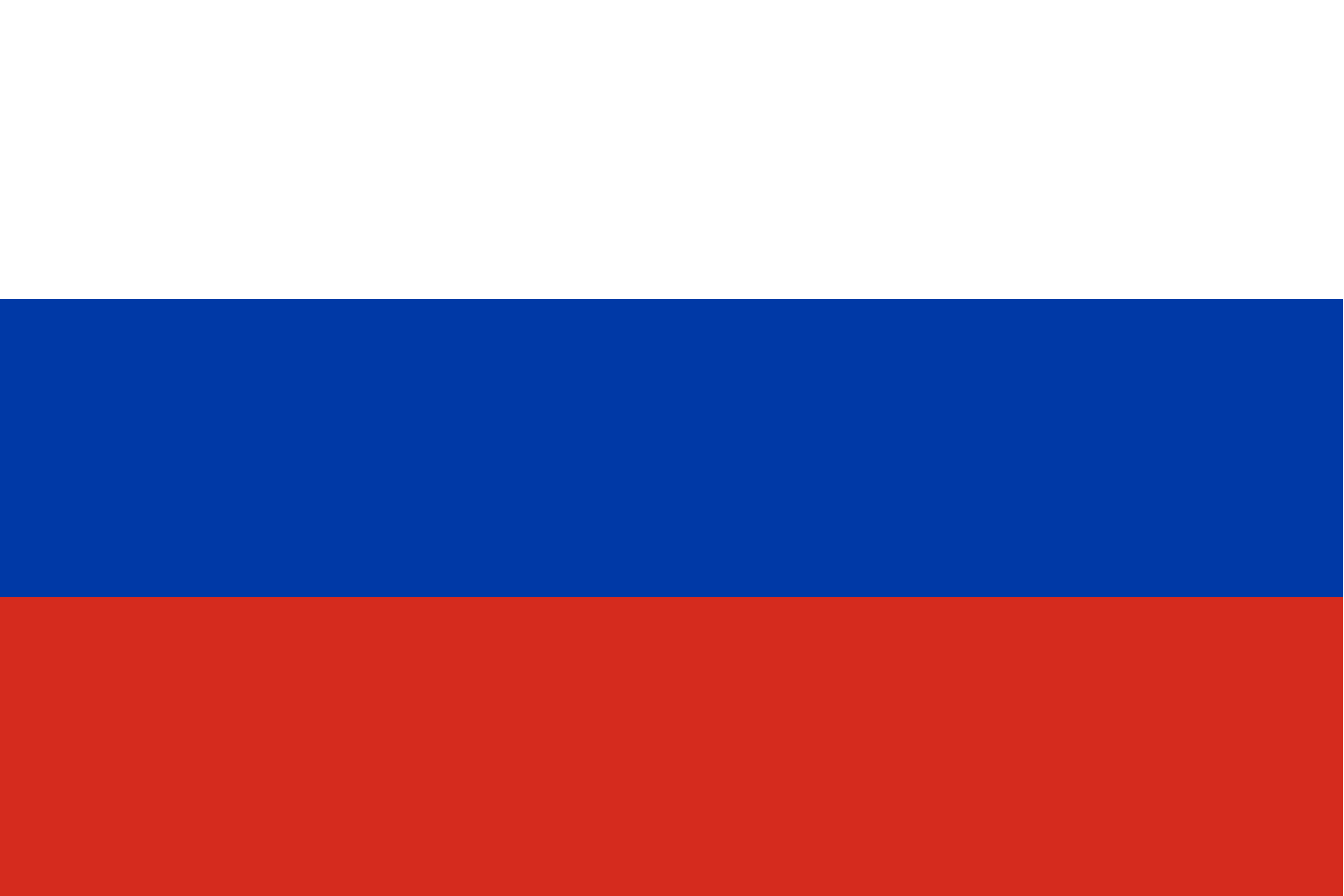}}}}
\newcommand{\UK}{{\setlength{\fboxsep}{0pt}\fbox{\includegraphics[height=0.30cm,width=0.45cm]{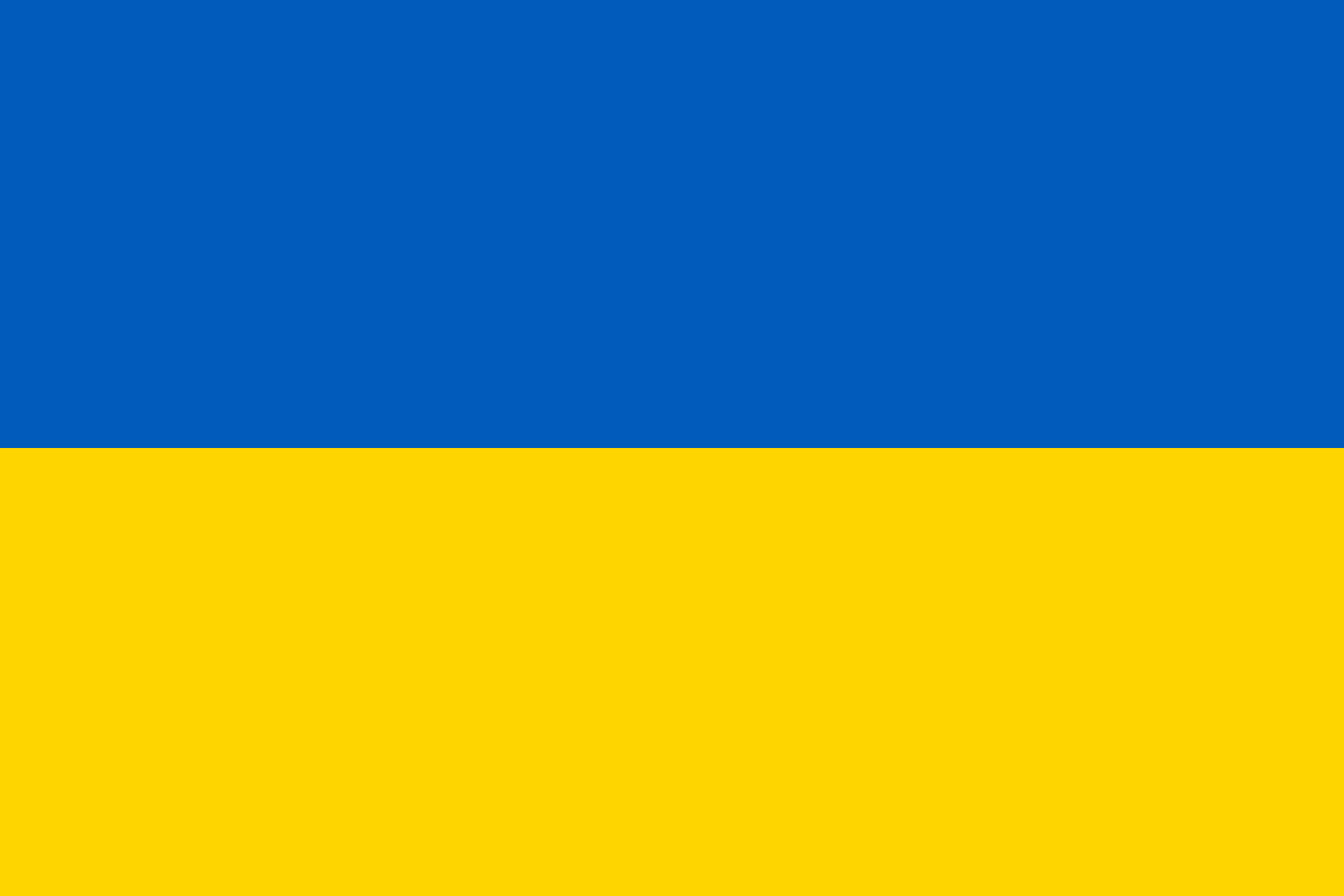}}}}
\newcommand{\CS}{{\setlength{\fboxsep}{0pt}\fbox{\includegraphics[height=0.30cm,width=0.45cm]{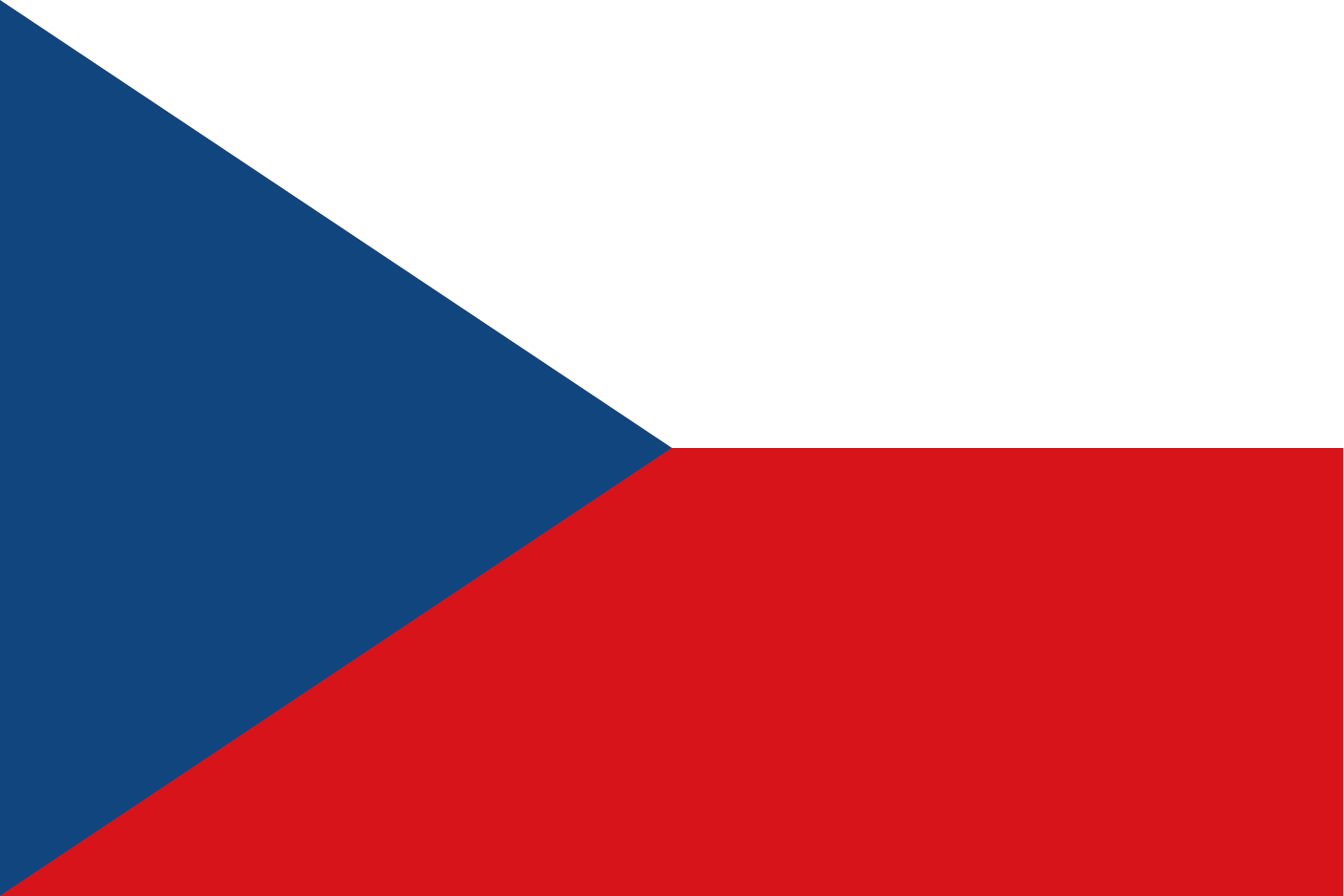}}}}
\newcommand{\BG}{{\setlength{\fboxsep}{0pt}\fbox{\includegraphics[height=0.30cm,width=0.45cm]{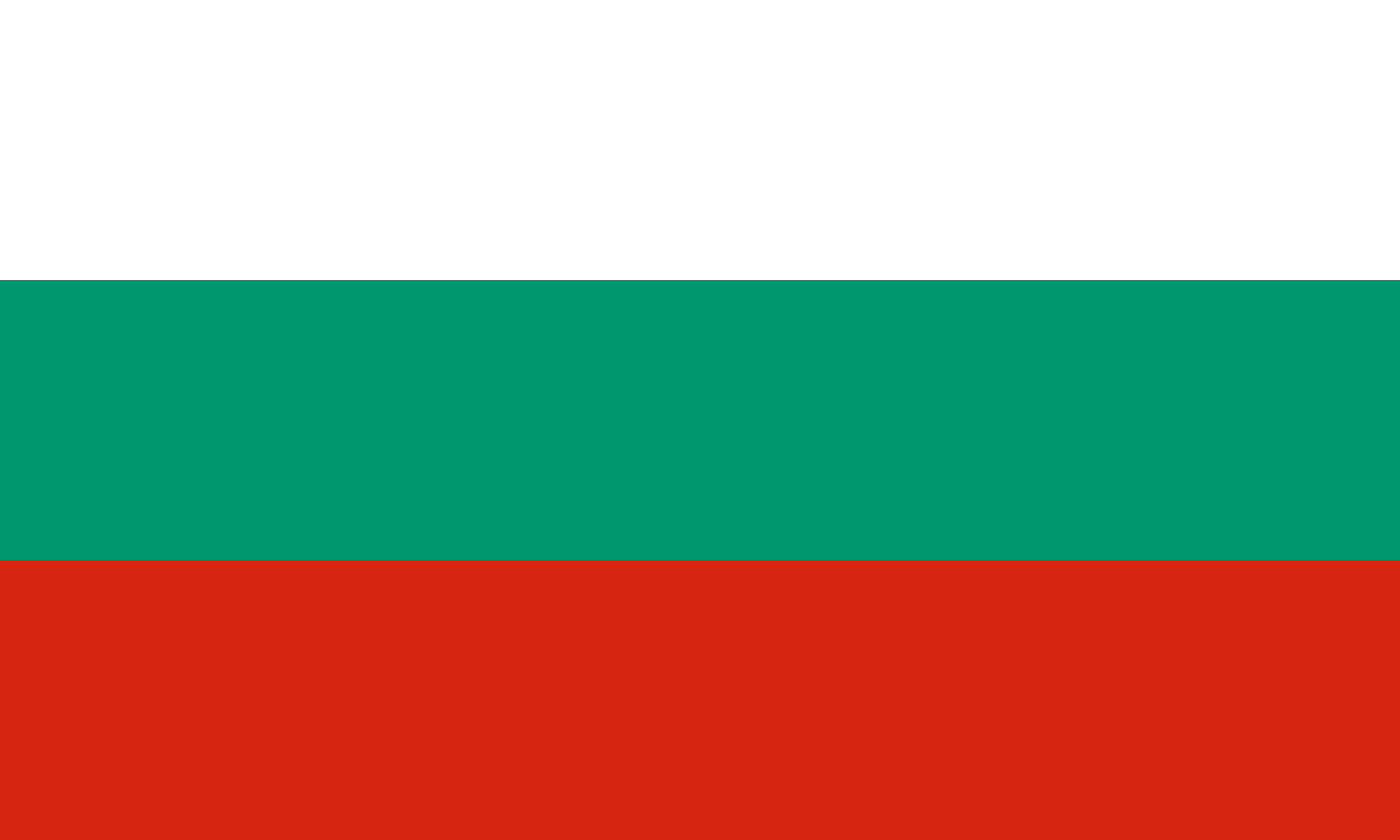}}}}
\newcommand{\SK}{{\setlength{\fboxsep}{0pt}\fbox{\includegraphics[height=0.30cm,width=0.45cm]{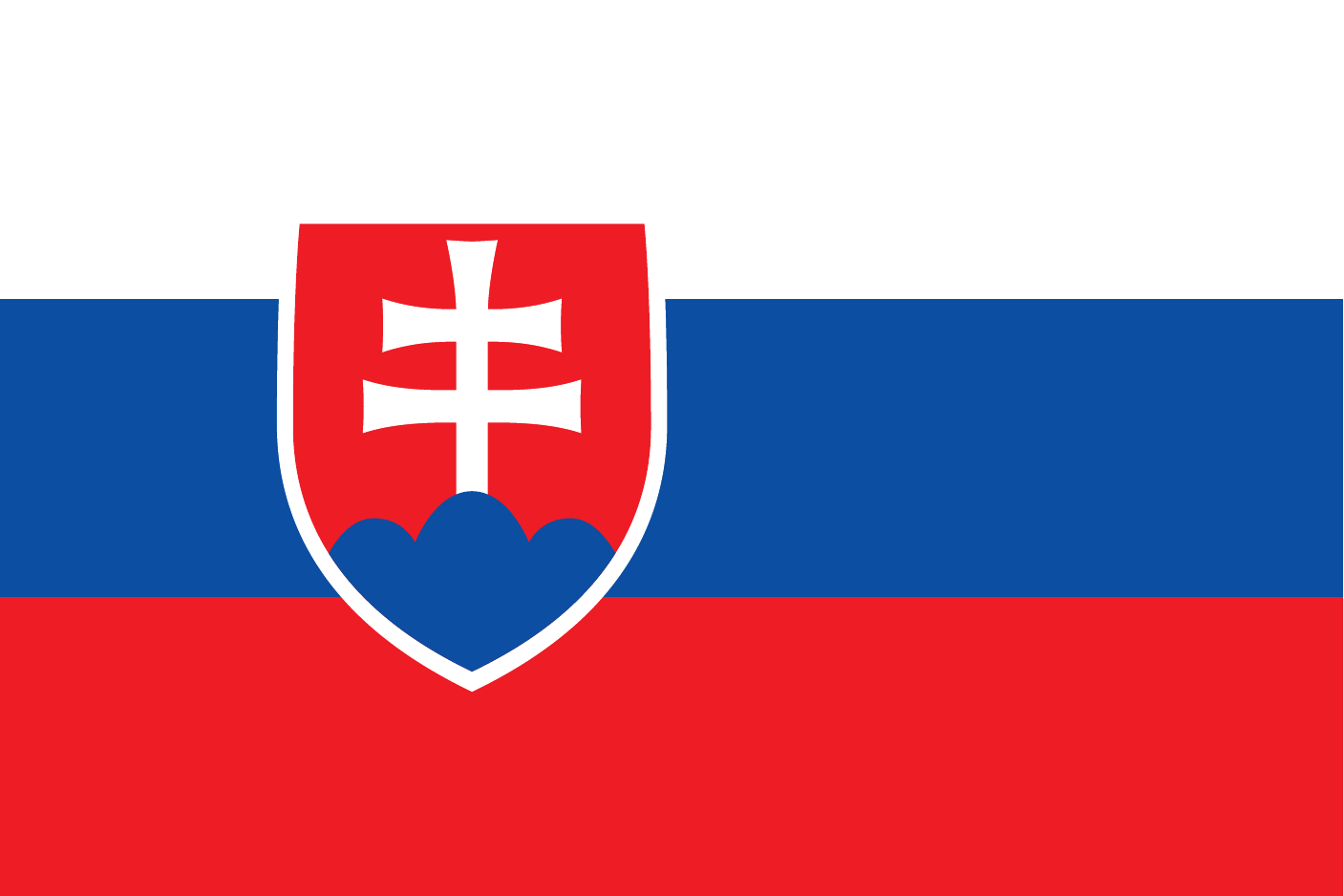}}}}
\newcommand{\CA}{{\setlength{\fboxsep}{0pt}\fbox{\includegraphics[height=0.30cm,width=0.45cm]{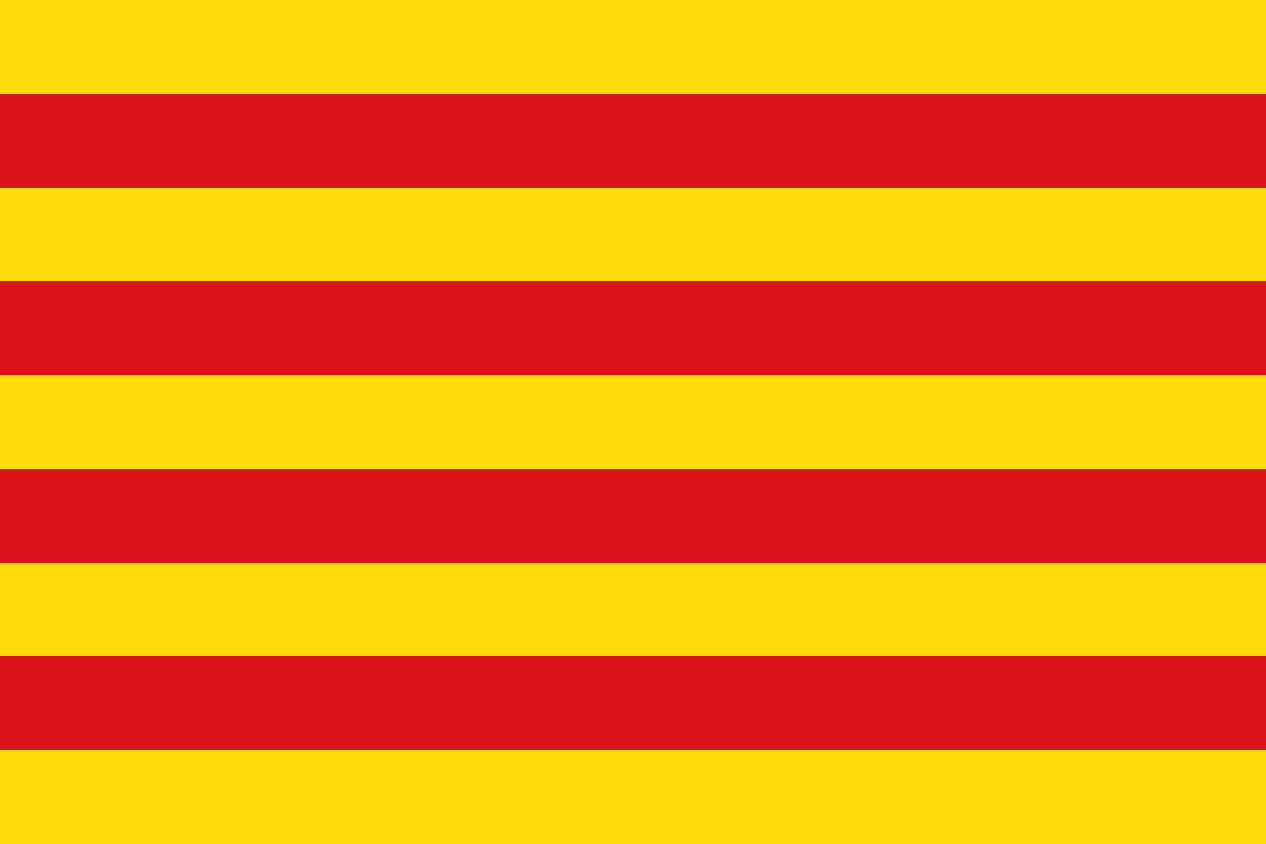}}}}
\newcommand{\ES}{{\setlength{\fboxsep}{0pt}\fbox{\includegraphics[height=0.30cm,width=0.45cm]{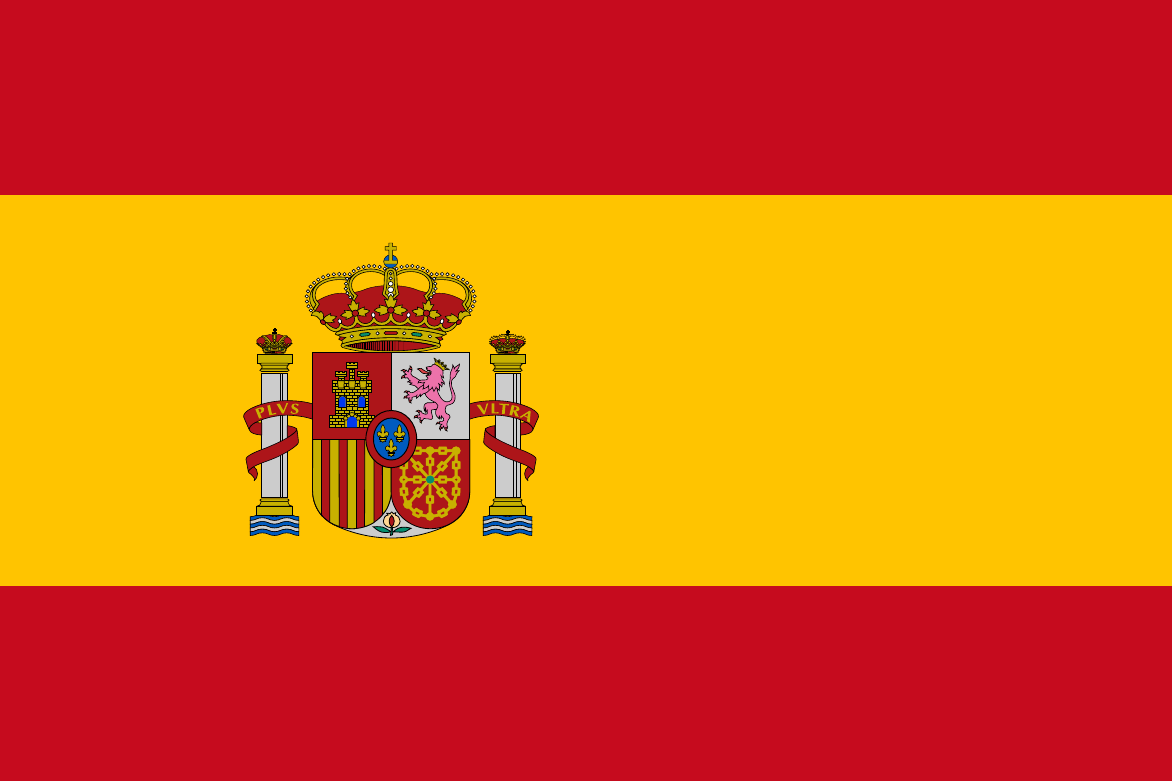}}}}
\newcommand{\PT}{{\setlength{\fboxsep}{0pt}\fbox{\includegraphics[height=0.30cm,width=0.45cm]{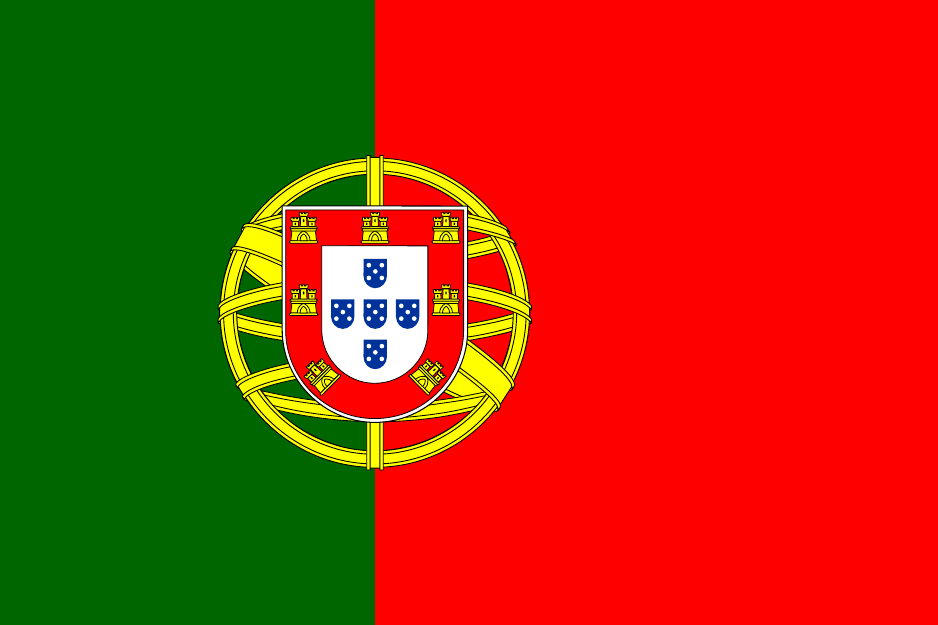}}}}
\newcommand{\IT}{{\setlength{\fboxsep}{0pt}\fbox{\includegraphics[height=0.30cm,width=0.45cm]{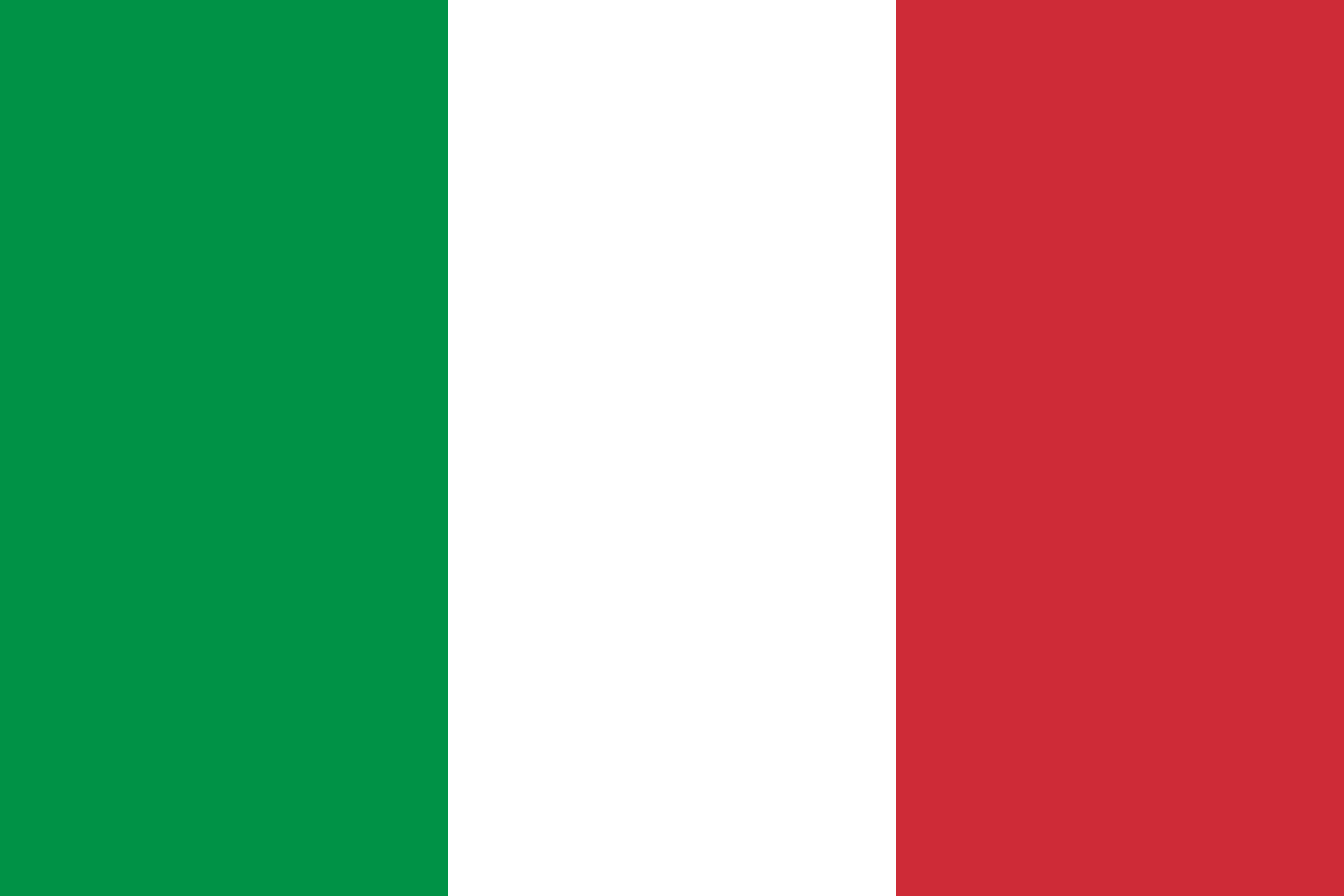}}}}
\newcommand{\FR}{{\setlength{\fboxsep}{0pt}\fbox{\includegraphics[height=0.30cm,width=0.45cm]{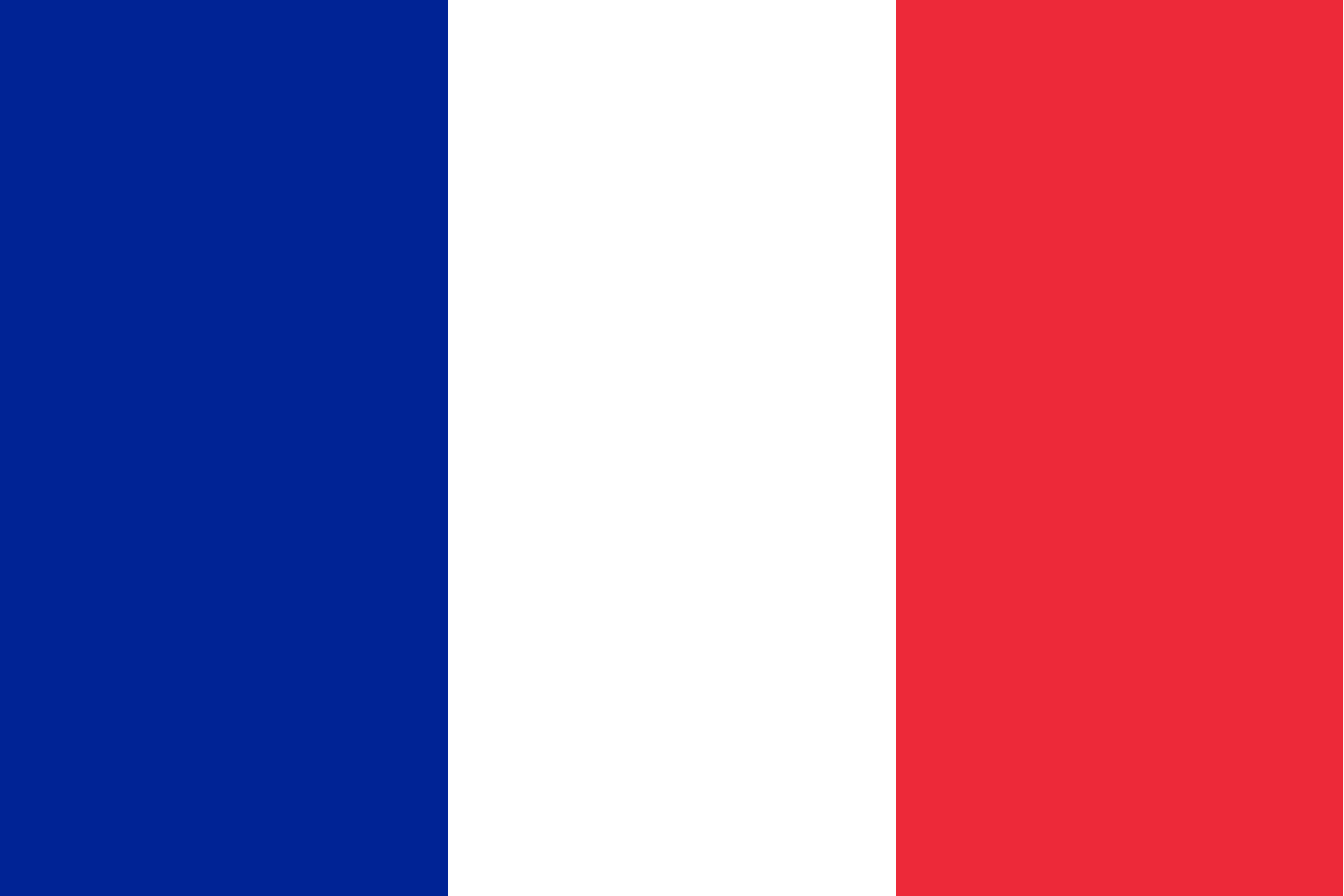}}}}
\newcommand{\RO}{{\setlength{\fboxsep}{0pt}\fbox{\includegraphics[height=0.30cm,width=0.45cm]{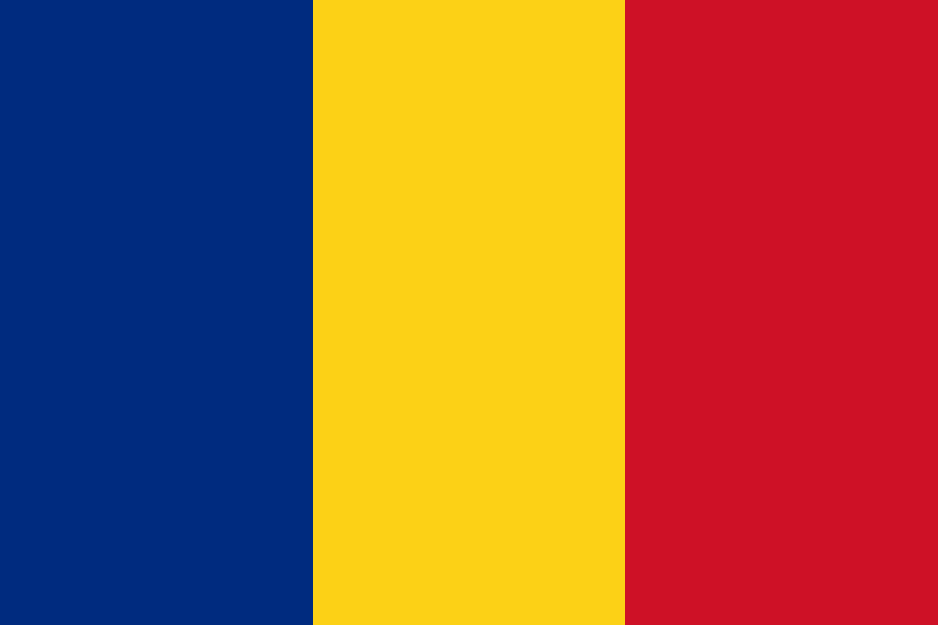}}}}
\newcommand{\SV}{{\setlength{\fboxsep}{0pt}\fbox{\includegraphics[height=0.30cm,width=0.45cm]{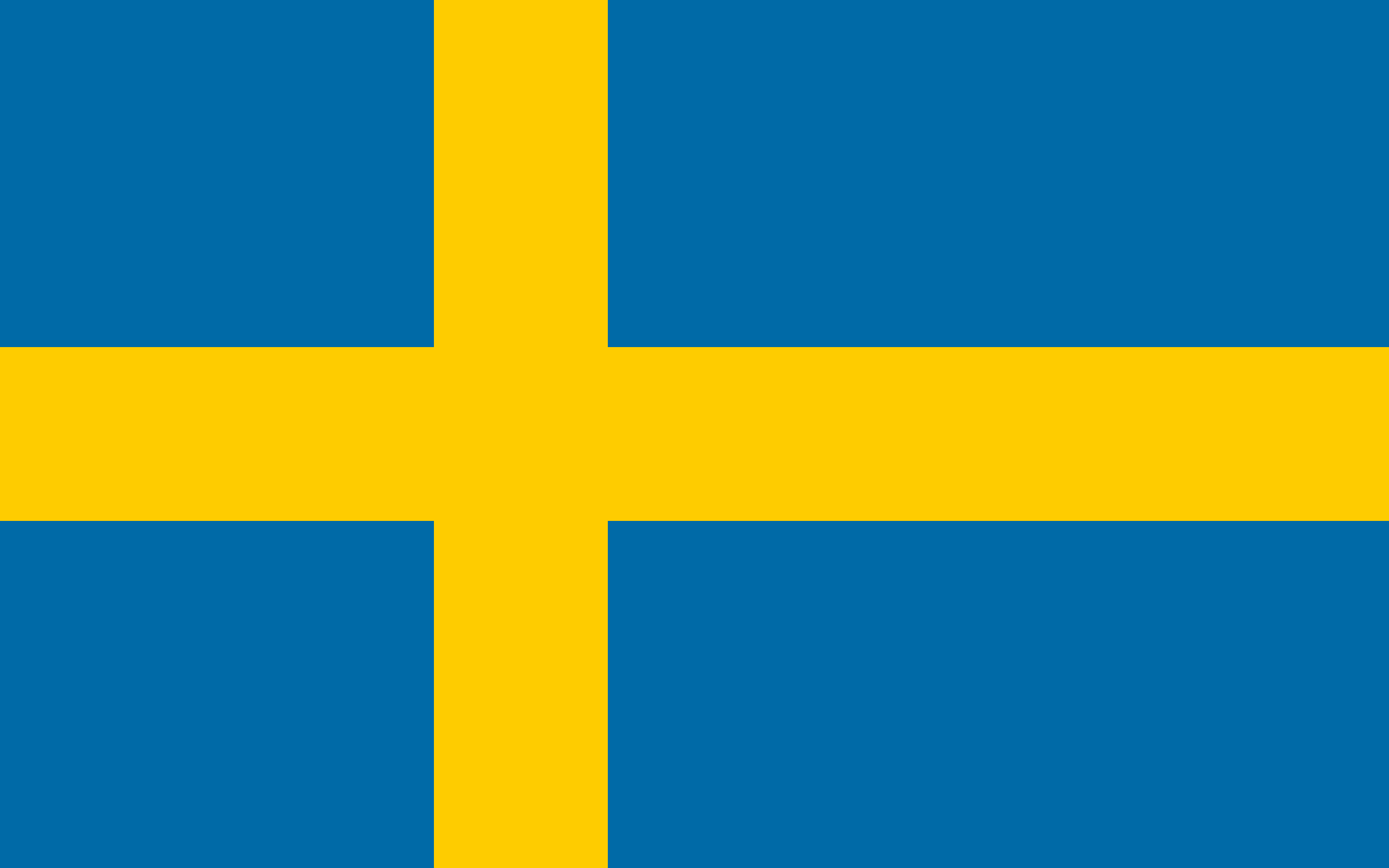}}}}
\newcommand{\NO}{{\setlength{\fboxsep}{0pt}\fbox{\includegraphics[height=0.30cm,width=0.45cm]{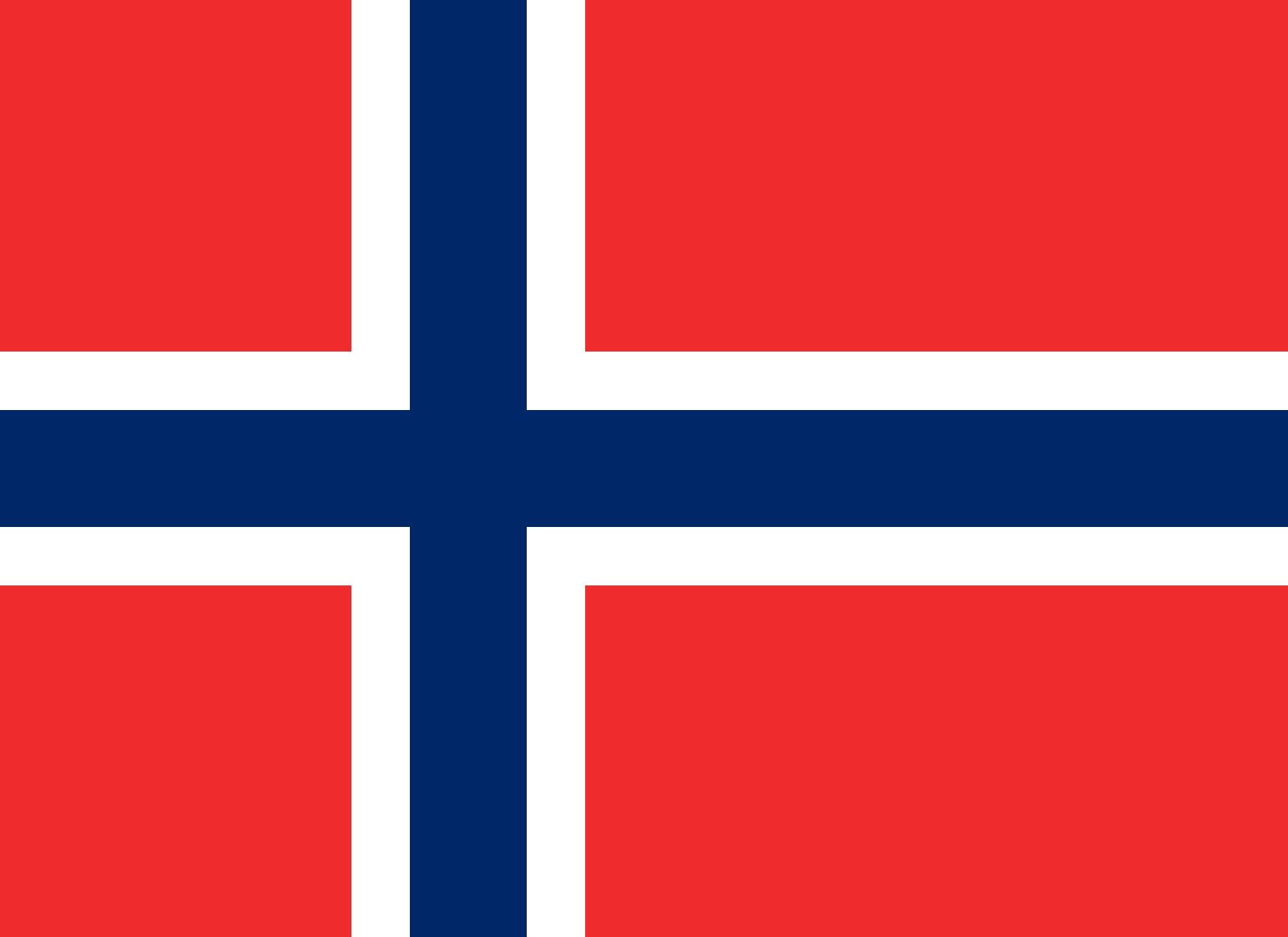}}}}
\newcommand{\DA}{{\setlength{\fboxsep}{0pt}\fbox{\includegraphics[height=0.30cm,width=0.45cm]{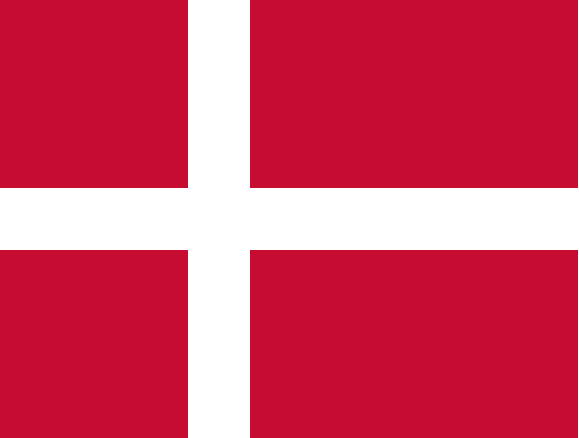}}}}
\newcommand{\HU}{{\setlength{\fboxsep}{0pt}\fbox{\includegraphics[height=0.30cm,width=0.45cm]{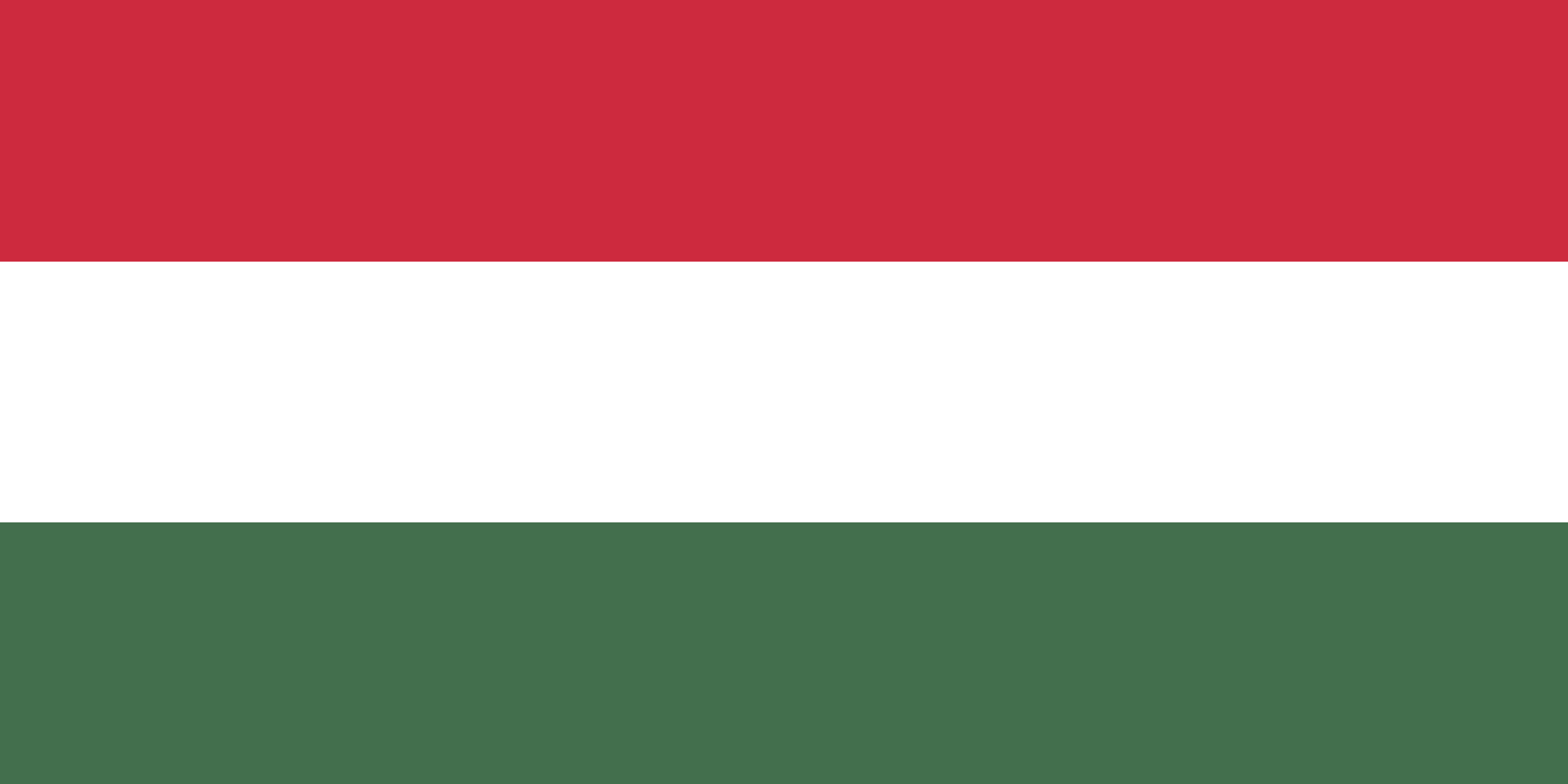}}}}
\newcommand{\FI}{{\setlength{\fboxsep}{0pt}\fbox{\includegraphics[height=0.30cm,width=0.45cm]{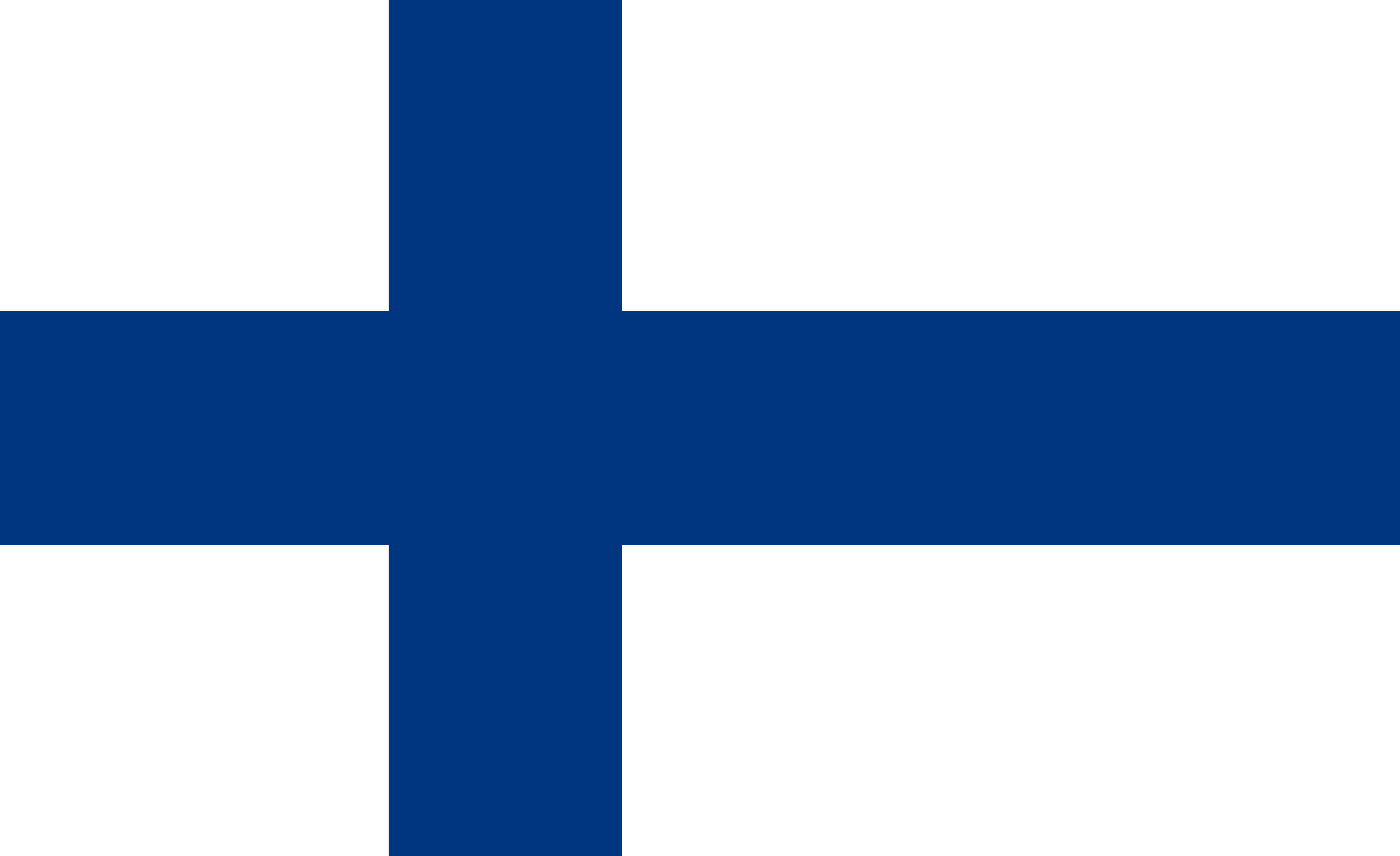}}}}
\newcommand{\ET}{{\setlength{\fboxsep}{0pt}\fbox{\includegraphics[height=0.30cm,width=0.45cm]{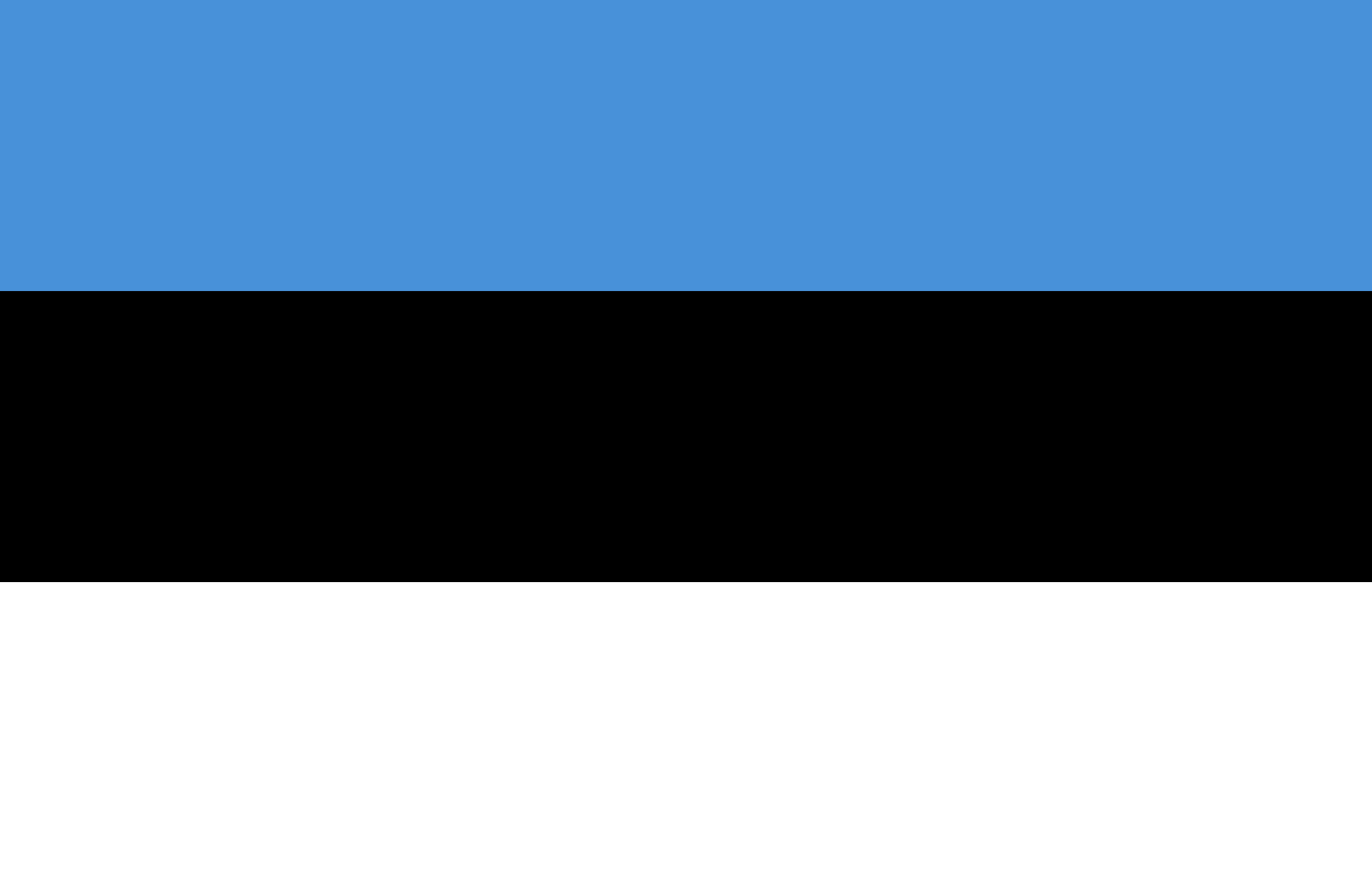}}}}

\newcommand{\vv}{\mathbf{v}}
\newcommand{\ve}{\mathbf{e}}
\newcommand{\vw}{{\boldsymbol w}}
\newcommand{\vt}{{\boldsymbol t}}
\newcommand{\vb}{\mathbf{b}}
\newcommand{\vtheta}{{\boldsymbol \theta}}

\title{Cross-lingual, Character-Level Neural Morphological Tagging} 
\author{
\textbf{Ryan Cotterell}\raise1.0ex\hbox{{\normalsize\textnormal  \textschwa}} \qquad
\textbf{Georg Heigold}\raise1.0ex\hbox{\normalsize\textnormal{\textipa{K}}}
 \\
\raise1.0ex\hbox{\normalsize \textschwa}Department of Computer Science, Johns Hopkins University, USA \\
\raise1.0ex\hbox{\normalsize \textipa{K}}German Research Center for Artificial Intelligence, Saarbr{\"u}cken, SL, GER \\
\texttt{\href{mailto:ryan.cotterell@jhu.edu}{\texttt{ryan.cotterell@jhu.edu}}} \qquad \texttt{\href{mailto:georg.heigold@gmail.com}{\texttt{georg.heigold@gmail.com}}}
}

\begin{document}
\maketitle
\begin{abstract}
  Even for common NLP tasks, sufficient
  supervision is not available in many languages---morphological
  tagging is no exception. In the work presented here, we explore
  a transfer learning scheme, whereby we
  train character-level recurrent neural taggers
  to predict morphological taggings for high-resource
  languages and low-resource languages together. Learning
  joint character representations among multiple related languages
  successfully enables knowledge transfer from the high-resource languages to the low-resource
  ones, improving accuracy by up to 30\% over a monolingual model.\looseness=-1
\end{abstract}

\section{Introduction}
State-of-the-art morphological taggers require thousands of annotated
sentences to train. For the majority of the world's languages,
however, sufficient, large-scale annotation is not available and
obtaining it would often be infeasible. Accordingly, an important road
forward in low-resource NLP is the development of methods that allow
for the training of high-quality tools from smaller amounts of
data. In this work, we focus on transfer learning---we train a
recurrent neural tagger for a low-resource language jointly with a
tagger for a related high-resource language. Forcing the models to
share character-level features among the languages allows large gains
in accuracy when tagging the low-resource languages, while maintaining
(or even improving) accuracy on the high-resource language.

Recurrent neural networks constitute the state of the art for a myriad
of tasks in NLP, e.g., multi-lingual
part-of-speech tagging \cite{plank-sogaard-goldberg:2016:P16-2},
syntactic parsing \cite{dyer-EtAl:2015:ACL-IJCNLP,zeman-EtAl:2017:K17-3}, morphological
paradigm completion \cite{cotterell-EtAl:2016:SIGMORPHON,cotterell-conll-sigmorphon2017} and language
modeling \cite{DBLP:conf/interspeech/SundermeyerSN12,melis2017state}; recently, such
models have also improved morphological tagging \cite{DBLP:journals/corr/HeigoldNG16,heigold2017}. In
addition to increased performance over classical approaches, neural
networks also offer a second advantage: they admit a clean paradigm
for multi-task learning.  If the learned representations for all of
the tasks are embedded jointly into a shared representation space, the various
tasks reap benefits from each other and often performance improves for
all \cite{DBLP:journals/jmlr/CollobertWBKKK11}.  We exploit this idea
for language-to-language transfer to develop an
approach for cross-lingual morphological tagging.\looseness=-1

We experiment on 18 languages taken from four different language
families. Using the Universal Dependencies treebanks, we emulate a
low-resource setting for our experiments, e.g., we attempt to train a
morphological tagger for Catalan using primarily data from a related
language like Spanish. Our results demonstrate the successful transfer
of morphological knowledge from the high-resource languages to the
low-resource languages without relying on an externally acquired
bilingual lexicon or bitext. We consider both the single- and
multi-source transfer case and explore how similar two languages must
be in order to enable high-quality transfer of morphological taggers.\footnote{While
  we only experiment with languages in the same family, we show that closer
  languages within that family are better candidates for transfer. We remark that future
  work should consider the viability of more distant language pairs.}

\section{Morphological Tagging}\label{sec:morpho-tagging}

\begin{figure*}
  \includegraphics[width=1.0\textwidth]{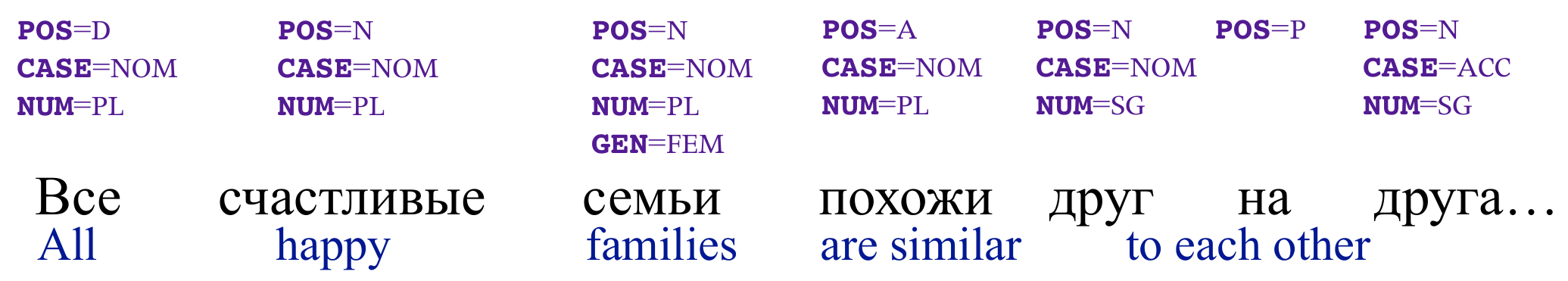}
  \caption{Example of a morphologically tagged sentence in Russian using the annotation
  scheme provided in the UD dataset.}
  \label{fig:russian-sentence}
\end{figure*}

Many languages in the world exhibit rich inflectional morphology: the
form of individual words mutates to reflect the syntactic
function. For example, the Spanish verb \word{so\~{n}ar} will appear
as \word{sue\~{n}o} in the first person present singular, but
\word{so\~{n}{\'a}is} in the second person present plural, depending
on the bundle of syntacto-semantic attributes associated with the given
form (in a sentential context). For concreteness, we list a more
complete table of Spanish verbal inflections in
\cref{tab:paradigm}.
Note that
some languages, e.g., the Northeast Caucasian language Archi, display a veritable
cornucopia of potential forms with the size of the verbal paradigm
exceeding 10,000 \cite{archi}.

Standard NLP annotation, e.g., the scheme in \newcite{sylakglassman-EtAl:2015:ACL-IJCNLP},
marks forms in terms of {\em universal} key--attribute pairs, e.g., the
first person present singular is represented as {\small
  $\left[\right.$\att{pos}{V}, \att{per}{1}, \att{num}{sg},
    \att{tns}{pres}$\left.\right]$}. 
This bundle of key--attribute
pairs is typically termed a morphological tag and we may view the goal of morphological tagging to label each
word in its sentential context with the appropriate tag \cite{oflazer1994tagging,hajic-hladka:1998:ACLCOLING}. As the
part of speech (POS) is a component of the tag, we may view
morphological tagging as a strict generalization of POS tagging, where we
have significantly refined the set of available tags. All of the experiments in this paper make use of the universal morphological tag set available
in the Universal Dependencies (UD) \cite{nivre2016universal}.  
As an example, we have
provided a Russian sentence with its UD tagging in
\cref{fig:russian-sentence}.

\begin{table}
\small
\centering
\begin{tabular}{lllll} \toprule
  & \multicolumn{2}{c}{{\sc present indicative}} & \multicolumn{2}{c}{{\sc past indicative}}\\ \midrule
      & \multicolumn{1}{c}{{\sc singular}} & \multicolumn{1}{c}{{\sc plural}} & \multicolumn{1}{c}{{\sc singular}} & \multicolumn{1}{c}{{\sc plural}}  \\ \cmidrule(r){2-3} \cmidrule(r){4-5}
  {\sc 1} & \word{sue\~{n}o}         & \word{{so\~{n}amos}} & \word{so\~{n}\'{e}} & \word{so\~{n}amos} \\
  {\sc 2} & \word{sue\~{n}as} & \word{so\~{n}\'{a}is}  & \word{so\~{n}aste} & \word{so\~{n}asteis}  \\
  {\sc 3} & \word{sue\~{n}a}  & \word{sue\~{n}an} & \word{so\~{n}\'{o}} & \word{so\~{n}aron} \\ \bottomrule
\end{tabular} 
  \caption{Partial inflection table for the Spanish verb
    \word{so\~{n}ar}.\looseness=-1}
  \label{tab:paradigm}
\end{table}

\paragraph{Transferring Morphology.}\label{sec:transferring-morphology}
The transfer of morphology is arguably more dependent on the relatedness of the languages in question than other linguistic annotation in NLP such as POS and named entity recognition (NER). 
For example, POS lends itself
nicely to a universal annotation scheme
\cite{DBLP:conf/lrec/PetrovDM12} and traditional NER is limited to a small number of cross-linguistic compliant categories, e.g., {\sc person} and {\sc place}.
Even universal dependency arc labels employ cross-lingual labels \cite{nivre2016universal}.
Morphology, on the other hand, typically requires more fine-grained annotation, e.g., grammatical case and tense. 
It is often the case
that one language will make a syntacto-semantic distinction in the form (or at
all) that another does not. 
For example, the Hungarian noun overtly marks 17
grammatical cases and Slavic verbs typically distinguish two aspects through morphology, while English marks none of these distinctions. 
If the word form in the source language does not overtly mark a grammatical category in the target language, it is
nigh-impossible to expect a successful transfer. 
For this reason, much of our work focuses on the transfer of related languages---specifically exploring \emph{how} close two languages must be for a successful transfer. 
Note that the language-specific nature of morphology does not contradict the universality of the annotation;
each language may mark a different subset of categories, i.e., use
a different set of the universal key--attribute pairs, but nevertheless there is a single, universal set, from which the key--attribute pairs are drawn.
See \newcite{newmeyer2007linguistic} for a linguistic treatment of cross-lingual annotation.

\paragraph{Notation.}
We will discuss morphological tagging in terms of the following notation. 
We will consider two (related)
languages: a high-resource \emph{source} language $\ell_s$ and a low-resource
{\em target} language $\ell_t$. 
Each of these languages will have its own (potentially
overlapping) set of morphological tags, denoted ${\cal T}_s$ and
${\cal T}_t$, respectively. 
We will work with the union
of both sets ${\cal T} = {\cal T}_s \cup {\cal T}_t$. 
An individual tag $t = \left[k_1\!\!=\!\!v_1,
  \ldots, k_J\!\!=\!\!v_J \right] \in {\cal T}$ is comprised of
universal keys and attributes, i.e., the pairs $(k_j, v_j)$ are completely
language-agnostic.
In the case where a language
does not mark a distinction, e.g., case on English nouns,
the corresponding keys are excluded from the tag. Typically, $|{\cal T}|$ is large (see \cref{tab:num-tags}).
We denote the set of training sentences for the high-resource source language
as ${\cal D}_s$ and the set of training sentences for the low-resource target language
as ${\cal D}_t$. 
In the experimental section, we will also consider a multi-source setting where we have multiple high-resource
languages, but, for ease of explication, we stick to the single-source
case in the development
of the model.

\begin{figure*}
  \begin{subfigure}[t]{0.57\columnwidth}
    \includegraphics[width=4.5cm]{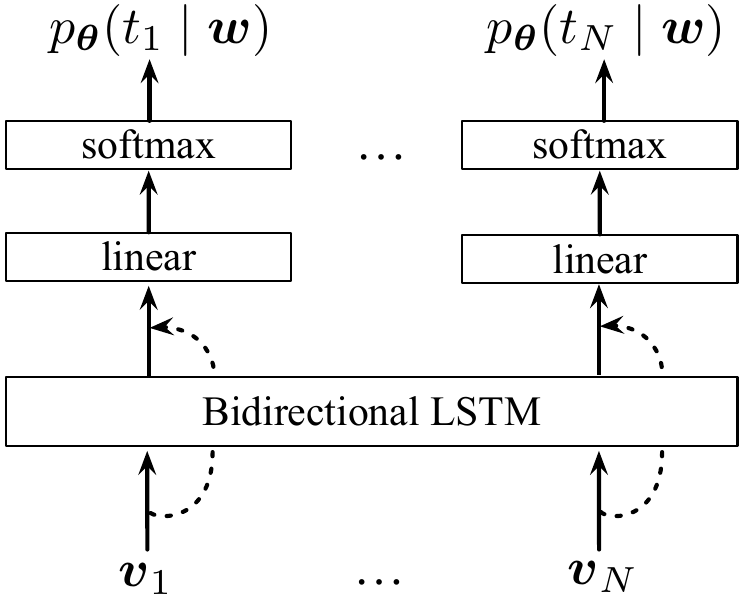}
    \caption{Vanilla architecture for neural morphological tagging.}
  \end{subfigure}
  ~
  \begin{subfigure}[t]{0.57\columnwidth}
    \includegraphics[width=4.5cm]{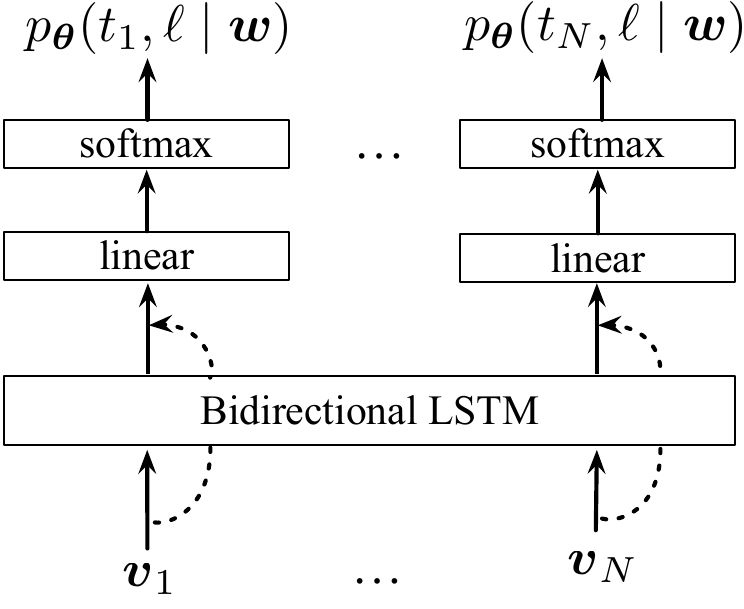}
    \caption{Joint morphological tagging and language identification.}
  \end{subfigure}
  ~
  \begin{subfigure}[t]{0.39\columnwidth}
    \includegraphics[height=3.5cm]{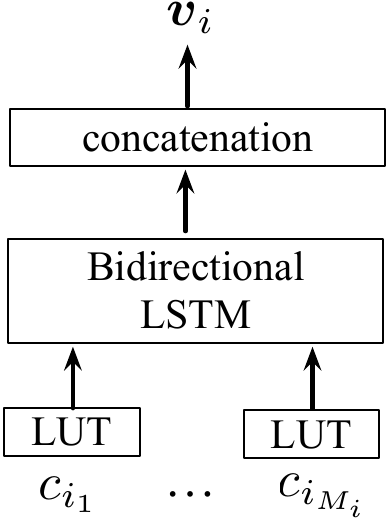}
    \caption{Character-level Bi-LSTM embedder.}
  \end{subfigure}
  ~
  \begin{subfigure}[t]{0.39\columnwidth}
    \includegraphics[height=3.5cm]{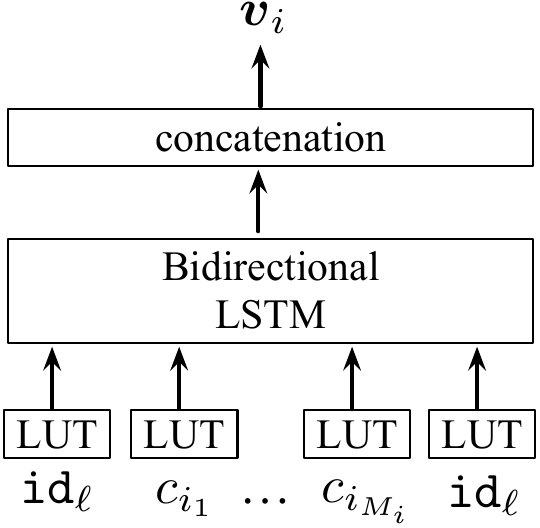}
    \caption{Language-specific Bi-LSTM embedder.}
  \end{subfigure}
  \label{fig:architectures}
  \caption{We depict four subarchitectures used in the models we develop in this work. Combining
    (a) with the character representations in (c) gives the vanilla morphological tagging architecture of \newcite{heigold2017}.
    Combining (a) with (d) yields the language-universal softmax architecture and (b) with (c) yields
    our joint model for language identification and tagging.}
\end{figure*}

\section{Character-Level Neural Transfer}
Our formulation of transfer learning builds on work in multi-task learning
\cite{DBLP:journals/ml/Caruana97,DBLP:journals/jmlr/CollobertWBKKK11}.
We treat each individual language as a task and train a joint model for all the tasks together.  
We first discuss the current state of the art
in morphological tagging: a character-level recurrent neural network. 
After that, we explore three augmentations to the
architecture that allow for the transfer learning scenario. 
All of our proposals force the representation of the characters for both the source and
the target language to share the same representation space, but involve
different mechanisms, by which the model may learn
language-specific features.

\subsection{Character-Level Neural Networks}\label{sec:character-level}
Character-level neural networks currently constitute the state
of the art in morphological tagging \cite{heigold2017}. We draw on previous work in defining a conditional
distribution over taggings $\vt$ for a sentence $\vw$ of length $|\vw| = N$ as
\begin{equation}
  p_{\vtheta}({\vt} \mid {\vw}) = \prod_{n=1}^N p_{\vtheta}(t_n \mid {\vw}), \label{eq:factorization}
\end{equation}
which may be seen as a $0^\text{th}$ order conditional random field (CRF)
\cite{DBLP:conf/icml/LaffertyMP01} with parameter vector ${\vtheta}$.\footnote{The parameter
  vector $\vtheta$ is a vectorization of all the parameters discussed below.} 
  Importantly, this factorization of
the distribution $p_{\vtheta}({\vt} \mid {\vw})$ also allows for efficient
exact decoding and marginal inference in ${\cal
  O}(N)$-time, but at the cost of not admitting any explicit interactions
in the output structure, i.e., between adjacent tags.\footnote{As an
  aside, it is quite interesting that a model with the factorization
  in \cref{eq:factorization} outperforms the {\sc MarMoT} model
  \cite{mueller-schmid-schutze:2013:EMNLP}, which focused on modeling
  higher-order interactions between the morphological tags, e.g., they employ
  up to a (pruned) $3^\text{rd}$ order CRF. That such a
  model achieves state-of-the-art performance indicates, however, that
  richer source-side features, e.g., those extracted by our
  character-level neural architecture, are more important for
  morphological tagging than higher-order tag interactions, which come
  with the added unpleasantness of exponential (in the order) decoding.\looseness=-1}
We parameterize the distribution over tags at each time step as
\begin{equation}
  p_{\vtheta}(t_n \mid {\vw}) = \mathrm{softmax}\left(\mathbf{W} \ve_n + \vb
  \right)_{t_n}, \label{eq:tagger}
\end{equation}
where $\mathbf{W} \in \mathbb{R}^{|{\cal T}| \times D}$ is a parameter matrix, $\vb \in
\mathbb{R}^{|{\cal T}|}$ is a bias vector and positional representations $\ve_n \in \mathbb{R}^D$ are
taken from a concatenation of the output of two long short-term memory recurrent neural networks (LSTMs) \cite{hochreiter1997long}, folded forward and
backward, respectively, over a sequence of input vectors.\footnote{See \cref{sec:details} for the exact values used in the experimentation.}  
The integer $D$ is the dimensionality and a tunable hyperparameter.
This constitutes a bidirectional LSTM \cite{DBLP:journals/nn/GravesS05}. 
We define
the positional representation vector as follows
\begin{align}
  \ve_n = &\left[ {\text{LSTM}}\left(\langle \vv_1, \ldots, 
    \vv_n\rangle \right); \right. \\
    &\,\,\,\,\,\,\,\,\,\,\,\,\,\,\,\,\,\,\,\,\,\,\,\,\,\,\,\,\,\,\,\,\,\,\,\,\,\, \left. {\text{LSTM}} \left(\langle \vv_{N}, \ldots, \vv_{n}\rangle \right) \right]. \nonumber \label{eq:embedder-e} 
\end{align}
where each $\vv_n \in \mathbb{R}^D$ is, itself, a word
representation.
Note that the function $\text{LSTM}$ returns
the {\em last} hidden state representation
of the network---a vector in $\mathbb{R}^{D/2}$, so that the concatenation of the
forward and backward states in \cref{eq:embedder-e} lies in $\mathbb{R}^{D}$
(and likewise for \cref{eq:embedder-v} below).
This architecture is the {\em context} bidirectional recurrent neural network
of \newcite{plank-sogaard-goldberg:2016:P16-2}. 
Finally, we derive each word representation $\vv_n$
from a character-level bidirectional LSTM embedder. 
Namely, we define each word representation
as the concatenation\looseness=-1
\begin{align}
  \vv_n = &\left[ {\text{LSTM}}\left(\langle c_{n_1}, \ldots, 
    c_{n_{M_n}}\rangle \right); \right. \\
    &\,\,\,\,\,\,\,\,\,\,\,\,\,\,\,\,\,\,\,\,\,\,\,\,\,\,\,\,\,\,\,\,\,\,\,\,\,\, \left. {\text{LSTM}} \left(\langle c_{n_{M_n}}, \ldots, c_{n_1}\rangle \right) \right]. \nonumber \label{eq:embedder-v} 
\end{align}
In other words, we run a bidirectional LSTM over the
character stream. This bidirectional LSTM is the {\em sequence} bidirectional recurrent neural network of \newcite{plank-sogaard-goldberg:2016:P16-2}. Note that a concatenation of the sequence of character symbols $\langle c_{n_1}, \ldots, c_{n_{M_n}} \rangle$ results in the word $w_n$.
Each of
the $M_n$ characters $c_{n_k}$ is a member of an alphabet $\Sigma$, the union of sets of characters in the languages considered.\looseness=-1

We direct the reader to
\newcite{heigold2017} for a more in-depth discussion of
this and various additional architectures for the computation of $\vv_n$; the
architecture we have presented in \cref{eq:embedder-v} is competitive
with the best-performing setting in \citeposs{heigold2017} study.

\subsection{Cross-Lingual Morphological Transfer as Multi-Task Learning}\label{sec:multi}
Cross-lingual morphological tagging may be formulated as a
multi-task learning problem. We seek to learn a set of shared
character representations for taggers in both languages together through
optimization of a joint loss function that combines the high-resource
tagger and the low-resource one. The first loss function
we consider is the following:
\begin{align}
  {\cal L}_{\textit{multi}}(\vtheta) =  -\!\!\!\sum_{(\vt, \vw) \in
    {\cal D}_s} \!\!\!\! \log&\,  p_{\vtheta} (\vt \mid \vw , \ell_s ) \\[-5\jot]
  \nonumber & -\!\!\!\!\sum_{(\vt, \vw) \in {\cal D}_t} \!\!
  \log p_{\vtheta}\left(\vt \mid \vw , \ell_t \right).
\end{align}
Crucially, our cross-lingual objective forces both taggers to share
part of the parameter vector $\vtheta$, which allows it to represent
morphological regularities between the two languages in a common
representation space and, thus, enables transfer of knowledge. 
This is
no different from monolingual multi-task settings, e.g., jointly
training a chunker and a tagger for the transfer of syntactic information
\cite{DBLP:journals/jmlr/CollobertWBKKK11}.  We point out that, in
contrast to our approach, almost all multi-task transfer learning,
e.g., for dependency parsing \cite{DBLP:conf/aaai/GuoCYWL16}, has shared word-level representations rather
than character-level representations. See \cref{sec:related-work} for a
more complete discussion.

We consider two parameterizations of this distribution $p_{\vtheta}(t_n
\mid \vw, \ell)$. 
First, we modify the initial character-level LSTM
representation such that it also encodes the identity of the language.
Second, we modify the softmax layer, creating a
language-specific softmax.

\paragraph{Language-Universal Softmax.}\label{par:arch1}
Our first architecture has one softmax, as in \cref{eq:tagger}, over all
morphological tags in ${\cal T}$ (shared among all the languages).
To allow the architecture to encode morphological features
specific to one language, e.g., the third person present plural
ending in Spanish is {\em -an}, but {\em -{\~a}o} in Portuguese,
we modify the creation of the character-level representations.
Specifically, we augment the character alphabet $\Sigma$ with a
distinguished symbol that indicates the language: $\text{{\tt id}}_\ell$. We
then pre- and postpend this symbol to the character stream
for every word before feeding the characters into the bidirectional LSTM.
Thus, we arrive at the new {\em language-specific} word representations,
\begin{align}
  \vv^{\ell}_n = &\left[ {\text{LSTM}}\left(\langle \text{{\tt id}}_\ell, c_{n_1}, \ldots, 
    c_{n_{M_n}}, \text{{\tt id}}_\ell \rangle \right); \right. \\
    &\,\,\,\,\,\,\,\,\,\,\,\,\,\,\,\,\,\, \left. {\text{LSTM}} \left(\langle \text{{\tt id}}_\ell, c_{n_{M_n}}, \ldots, c_{n_1}, \text{{\tt id}}_\ell \rangle \right) \right]. \nonumber
\end{align}
This model creates a language-specific representation $\vv_n$, but the
individual representations for a given character are shared among
the languages jointly trained on. 
The remainder of the architecture is held constant.

\paragraph{Language-Specific Softmax.}\label{par:arch2}
Next, inspired by the architecture of \newcite{heigold2013multilingual}, we consider a language-specific
softmax layer, i.e., we define a new output layer for every
language:
\begin{equation}
  p_{\vtheta}\left(t_n \mid \vw, \ell \right) = \mathrm{softmax}\left(\mathbf{W}_{\ell} \ve_n + \vb_{\ell}\right)_{t_n},
  \label{eq:lang-specific}
\end{equation}
where $\mathbf{W}_{\ell} \in \mathbb{R}^{|{\cal T}| \times D}$ and $\vb_{\ell} \in \mathbb{R}^{|{\cal T}|}$ are now \emph{language-specific}.
In this architecture, the representations $\ve_n$ are the same for all
languages---the model has to learn language-specific behavior exclusively through the
output softmax of the tagging LSTM.\looseness=-1

\paragraph{Joint Morphological Tagging and Language Identification.}\label{sec:joint-arch}
The third model we exhibit is a joint architecture for tagging and
language identification. We consider the following loss function:
\begin{align}
  {\cal L}_{\textit{joint}} (\vtheta) =  -\!\!\!\sum_{(\vt, \vw) \in {\cal D}_s} \!\!\! \log\, & p_{\vtheta}(\ell_s, \vt \mid \vw) \\[-5\jot] \nonumber
  &-\!\sum_{(\vt, \vw) \in {\cal D}_t} \!\!\!\!\! \log p_{\vtheta}\left(\ell_t, \vt \mid \vw\right),
 \end{align}
where we factor the joint distribution as 
\begin{align}
  p_{\vtheta}\left(\ell, \vt \mid \vw \right) &= p_{\vtheta}\left(\ell \mid \vw \right) \cdot p_{\vtheta}\left(\vt \mid \vw, \ell \right). 
\end{align}
Just as before, we define $p_{\vtheta}\left(\vt \mid \vw, \ell \right)$ above as in \cref{eq:lang-specific} and
we define
\begin{equation}
  p_{\vtheta}(\ell \mid \vw) = \mathrm{softmax}\left(\mathbf{U}\tanh(\mathbf{V}\bar{\ve})\right)_{\ell},
\end{equation}
where $\bar{\ve} = \frac{1}{N}\sum_{n=1}^{N} \ve_n$ is a single sentence-level representation obtained by
mean-pooling the positional representations $\ve_n$ over the sentence.
This is a multi-layer perceptron with a binary softmax (over the two languages)
as an output layer; we have added the additional parameters $\mathbf{V} \in
\mathbb{R}^{H \times D}$ and $\mathbf{U} \in \mathbb{R}^{2 \times H}$, where $H$ is
the size of the hidden layer.
In the
case of multi-source transfer, the output softmax instead ranges over the \emph{set} of
languages, so $\mathbf{U}$ has one row per language rather than two.

\begin{table}
  \begin{adjustbox}{width=1.\columnwidth}
    \begin{tabular}{llll llll} \toprule
      \multicolumn{4}{c}{Romance} & \multicolumn{4}{c}{Slavic}  \\ \cmidrule(r){1-4} \cmidrule(r){5-8}
      lang & train & dev  & test & lang & train & dev  & test   \\ \midrule
      \CA\,(ca)   & 13123 & 1709 & 1846 & \BG\,(bg)   & 8907  & 1115 & 1116   \\
      \ES\,(es)   & 14187 & 1552 & 274  & \CS\,(cs)   & 61677 & 9270 & 10148  \\
      \FR\,(fr)   & 14554 & 1596 & 298  & \PL\,(pl)   & 6800  & 700  & 727    \\
      \IT\,(it)   & 12837 & 489  & 489  & \RU\,(ru)   & 4029  & 502  & 499    \\
      \PT\,(pt)   & 8800  & 271  & 288  & \SK\,(sk)   & 8483  & 1060 & 1061   \\
      \RO\,(ro)   & 7141  & 1191 & 1191 & \UK\,(uk)   & 200   & 30   & 25     \\ \toprule
      \multicolumn{4}{c}{Germanic} & \multicolumn{4}{c}{Uralic} \\ \cmidrule(r){1-4} \cmidrule(r){5-8}
      lang        & train & dev  & test & lang        & train & dev  & test   \\ \midrule
      \DA\,(da)   & 4868  & 322  & 322  & \ET\,(et)   & 14510 & 1793 & 1806   \\
      \NO\,(no)   & 15696 & 2410 & 1939 & \FI\,(fi)   & 12217 & 716  & 648    \\
      \SV\,(sv)   & 4303  & 504  & 1219 & \HU\,(hu)   & 1433  & 179  & 188    \\ \bottomrule
  \end{tabular}
  \end{adjustbox}
  \caption{Number of sentences in each of the train, development and test splits (organized by language family).}
  \label{tab:lang-size}
\end{table}

\paragraph{Comparative Discussion.}
The first two architectures discussed in \cref{par:arch1} represent two
possibilities for a multi-task objective, where we condition on the
language of the sentence. The first integrates this knowledge at a
lower level and the second at a higher level. The third architecture
discussed in \cref{sec:joint-arch} takes a different tack---rather
than conditioning on the language, it predicts it. The joint model
offers one interesting advantage over the two architectures
proposed. Namely, it allows us to perform a morphological analysis on
a sentence where the language is unknown. This effectively
alleviates an early step in the NLP pipeline, where language
id is performed, and is useful in conditions
where the language to be tagged may not be known {\em a priori},
e.g., when tagging social media data.
While there are certainly more complex architectures one could
engineer for the task, we believe we have found a relatively diverse
sampling, enabling an interesting experimental comparison. Indeed, it
is an important empirical question which architectures are most
appropriate for transfer learning. Since transfer learning affords the
opportunity to reduce the sample complexity of the ``data-hungry''
neural networks that currently dominate NLP research, finding a good
solution for cross-lingual transfer in state-of-the-art neural models
will likely be a boon for low-resource NLP in general.

\section{Experiments}\label{sec:experiments}
Empirically, we ask three questions of our
architectures. i) How well can we transfer morphological tagging
models from high-resource languages to low-resource languages in each architecture? (Does
one of the three outperform the others?) ii)
How much annotated data in the low-resource language do we need? iii)
How closely related do the languages need to be to get good transfer?

\subsection{Experimental Languages}
We experiment with the language families:
Romance (Indo-European), Northern Germanic (Indo-European), Slavic (Indo-European)
and Uralic.
In the Romance
sub-grouping of the wider Indo-European family, we experiment on Catalan (ca), French (fr), Italian (it),
Portuguese (pt), Romanian (ro) and Spanish (es). In the Northern Germanic
family, we experiment on Danish (da), Norwegian (no) and Swedish (sv).
In the Slavic family, we experiment on Bulgarian (bg), Czech (cs), Polish (pl),
Russian (ru), Slovak (sk) and Ukrainian (uk). Finally, in the Uralic
family we experiment on Estonian (et), Finnish (fi) and Hungarian (hu).
\begin{table}
  \begin{adjustbox}{width=1.\columnwidth}
  \begin{tabular}{llllllll} \toprule
    \multicolumn{2}{c}{Romance} & \multicolumn{2}{c}{Slavic} & \multicolumn{2}{c}{Germanic} & \multicolumn{2}{c}{Uralic} \\ \cmidrule(r){1-2} \cmidrule(r){3-4} \cmidrule(r){5-6} \cmidrule(r){7-8}
    lang & $|{\cal T}|$ & lang & $|{\cal T}|$ & lang & $|{\cal T}|$ & lang & $|{\cal T}|$ \\ \midrule
    \CA\,(ca) & 172  & \BG\,(bg) & 380  & \DA\,(da) & 124   & \ET\,(et) & 654 \\
    \ES\,(es) & 232  & \CS\,(cs) & 2282 & \NO\,(no) & 169   & \FI\,(fi) & 1440 \\
    \FR\,(fr) & 142  & \PL\,(pl) & 774  & \SV\,(sv) & 155   & \HU\,(hu) & 634\\
    \IT\,(it) & 179  & \RU\,(ru) & 520  \\
    \PT\,(pt) & 375  & \SK\,(sk) & 597  \\
    \RO\,(ro) & 367  & \UK\,(uk) & 220   \\ \bottomrule
  \end{tabular}
  \end{adjustbox}
  \caption{Number of unique morphological tags
    for each of the experimental languages (organized by family).\looseness=-1}
  \label{tab:num-tags}
\end{table}

\subsection{Datasets}
We use the morphological tagging datasets provided by the Universal Dependencies (UD) treebanks (the concatenation
of the $4^\text{th}$ and $6^\text{th}$ columns of the file
format)
\cite{nivre2016universal}. We list the size of the training,
development, and test splits of the UD treebanks we used in
\cref{tab:lang-size}. Also, we list the number of unique morphological
tags in each language in \cref{tab:num-tags}, which serves as an
approximate measure of the morphological complexity each language
exhibits. Crucially, the data are annotated in a cross-linguistically
consistent manner, such that words in the different languages that
have the same syntacto-semantic function have the same bundle of tags
(see \cref{sec:morpho-tagging} for a discussion). Potentially, further gains would be
possible by using a more universal scheme, e.g., the {\sc UniMorph} scheme.

\subsection{Baselines}\label{sec:baselines}
We consider two baselines in our work. First, we consider
the {\sc MarMoT} tagger \cite{mueller-schmid-schutze:2013:EMNLP},
which is currently the best-performing non-neural model.
The source code for {\sc MarMoT} is freely available online,\footnote{\url{http://cistern.cis.lmu.de/marmot/}}
which allows us to perform fully controlled experiments
with this model.
Second, we consider the alignment-based projection
approach of \newcite{buys-botha:2016:P16-1}.\footnote{We do
  not have access to the code as the model was developed in industry,
  so we compare to the numbers reported in the original paper, as well
  as additional numbers provided to us by the first author in a personal communication. The numbers will not be, strictly speaking, comparable. However,
  we hope they provide insight into the relative performance of the tagger.\looseness=-1} We
discuss each of the two baselines in turn. 

\begin{table*}[t]
  \centering
  \begin{subtable}[b]{2.0\columnwidth}
    \centering
    \begin{adjustbox}{width=1.\columnwidth}
      \begin{tabular}{clcccccccccccc} \toprule
  &  & \multicolumn{12}{c}{target language} \\
  & & \multicolumn{6}{c}{$|{\cal D}_t| = 100$} & \multicolumn{6}{c}{$|{\cal D}_t| = 1000$} \\ \cmidrule(r){3-8} \cmidrule(r){9-14}
&& \CA\,(ca) & \ES\,(es)      & \FR\,(fr)      & \IT\,(it)      & \PT\,(pt)      & \RO\,(ro)      & \CA\,(ca)      & \ES\,(es)      & \FR\,(fr)      & \IT\,(it)      & \PT\,(pt)      & \RO\,(ro)     \\ \midrule
        \multirow{6}{*}{\rotatebox{90}{source language}}
&  \CA\,(ca)           & ---    & 87.9\% & 84.2\% & 84.6\% & 81.1\% & 67.4\% & ---    & 94.1\% & {\bf 93.5\%} & 93.1\% & {\bf 89.0\%}  & 89.8\% \\
&  \ES\,(es)           & 88.9\% & ---    & 85.5\% & 85.6\% & 81.8\% & 69.5\% & {\bf 95.5\%} & ---    & {\bf 93.5\%} & 93.5\% & 88.9\%  & 89.7\% \\
&  \FR\,(fr)           & 88.3\% & 87.0\% & ---    & 83.6\% & 79.5\% & {\bf 69.9\%} & 95.4\% & 93.8\% & ---    & 93.3\% & 88.6\%  & 89.7\% \\
&  \IT\,(it)           & 88.4\% & 87.8\% & 84.2\% & ---    & 80.6\% & 69.1\% & 95.4\% & 94.0\% & 93.3\% & ---    & 88.7\%  & {\bf 90.3\%} \\
&  \PT\,(pt)           & 88.4\% & 88.9\% & 85.1\% & 84.7\% & ---    & 69.6\% & 95.3\% & {\bf 94.2\%} & {\bf 93.5\%} & 93.6\% & ---     & 89.8\% \\
&  \RO\,(ro)           & 87.6\% & 87.2\% & 85.0\% & 84.4\% & 79.9\% & ---    & 95.3\% & 93.6\% & 93.4\% & 93.2\% & 88.5\%  & --- \\ \midrule
        & multi-source & {\bf 89.8\%} & {\bf 90.9\%} & {\bf 86.6\%} & {\bf 86.8\%} & {\bf 83.4\%} & 67.5\% & 95.4\% & {\bf 94.2\%} & 93.4\% & {\bf 93.8\%}  & 88.7\% & 88.9\% \\ \bottomrule
  \end{tabular}
    \end{adjustbox}
    \caption{Results for the Romance languages.}
  \end{subtable}
  \par\bigskip
  \begin{subtable}[b]{2.0\columnwidth}
    \centering
    \begin{adjustbox}{width=1.0\columnwidth}
      \begin{tabular}{clcccccccccccccc} \toprule
     &   & \multicolumn{12}{c}{target language} \\
  & & \multicolumn{6}{c}{$|{\cal D}_t| = 100$} & \multicolumn{6}{c}{$|{\cal D}_t| = 1000$} \\ \cmidrule(r){3-8} \cmidrule(r){9-14}
   &    & \BG\,(bg)     & \CS\,(cs)     & \PL\,(pl)     & \RU\,(ru)     & \SK\,(sk)     & \UK\,(uk)     & \BG\,(bg)     & \CS\,(cs)     & \PL\,(pl)    & \RU\,(ru)     & \SK\,(sk)    & \UK\,(uk)  \\ \midrule
        \multirow{6}{*}{\rotatebox{90}{source language}}
    &\BG\,(bg)             & ---    & 47.4\% & 44.7\% & 67.3\% & 39.7\% & 57.3\% & ---    & 73.7\% & 75.0\% & 84.1\%  & 70.9\% & 72.0\%   \\
    &\CS\,(cs)             & 57.8\% & ---    & 56.5\% & 62.6\% & 62.6\% & 54.0\% & 80.9\% & ---    & {\bf 80.0\%} & 84.1\%  & 78.1\% & 64.7\%   \\
    &\PL\,(pl)             & 54.3\%          & 54.0\% & ---    & 59.3\% & 57.8\% & 48.0\% & 78.3\% & 74.9\% & ---    & {\bf 84.2\%}  & 75.9\% & 57.3\%  \\
    &\RU\,(ru)             & {\bf 68.8\%}    & 48.6\% & 47.4\% & ---    & 46.5\% & {\bf 60.7\%} & {\bf 83.1\%} & 73.6\% & 76.0\% & ---     & 71.4\% & {\bf 72.7\%}   \\
    &\SK\,(sk)             & 55.2\% & 57.4\% & 54.8\% & 61.2\% & ---    & 49.3\% & 77.6\% & {\bf 76.3\%} & 78.4\% & 83.9\%  & ---    & 60.7\%  \\
    &\UK\,(uk)             & 44.1\% & 36.0\% & 34.4\% & 43.2\% & 30.0\% & ---    & 67.3\% & 64.8\% & 66.9\% & 76.1\%  & 56.0\% & ---  \\ \midrule
   & multi-source& 64.5\% & {\bf 57.9\%} & {\bf 57.0\%} & {\bf 64.4\%} & {\bf 64.8\%} & 58.7\% & 81.6\% & 74.8\% & 78.1\% & 83.1\%  & {\bf 79.6\%} & 69.3\% \\ \bottomrule
  \end{tabular}
    \end{adjustbox}
    \caption{Results for the Slavic languages.}
  \end{subtable}
  \par\bigskip
  \begin{subtable}[b]{1.0\columnwidth}
    \centering
    \begin{adjustbox}{width=1.\columnwidth}
      \begin{tabular}{clcccccc} \toprule
   &     & \multicolumn{6}{c}{target language} \\
  & & \multicolumn{3}{c}{$|{\cal D}_t| = 100$} & \multicolumn{3}{c}{$|{\cal D}_t| = 1000$} \\ \cmidrule(r){3-5} \cmidrule(r){6-8}
&     & \DA\,(da)      & \NO\,(no)      & \SV\,(sv)      & \DA\,(da)      & \NO\,(no)      & \SV\,(sv)      \\ \midrule
        \multirow{3}{*}{\rotatebox{90}{source}}
  &  \DA\,(da)           & ---     & 77.6\%  & 73.1\%  & ---     & 90.1\%  & 90.0\%  \\
  &  \NO\,(no)           & 83.1\%  & ---     & 75.7\%  & 93.1\%  & ---     & 90.5\%  \\
  &  \SV\,(sv)           & 81.4\%  & 76.5\%  & ---     & 92.6\%  & 90.2\%  & ---  \\ \midrule
& multi-source & {\bf 87.8\%} & {\bf 82.3\%} & {\bf 77.2\%}  & {\bf 93.9\%}  & {\bf 91.2\%} & {\bf 90.9\%} \\ \bottomrule
    \end{tabular}
    \end{adjustbox}
    \caption{Results for the Northern Germanic languages.}
  \end{subtable}
  ~
    \begin{subtable}[b]{1.0\columnwidth}
    \centering
   \begin{adjustbox}{width=1.\columnwidth}
     \begin{tabular}{clcccccc} \toprule
    &   & \multicolumn{6}{c}{target language} \\
 &  & \multicolumn{3}{c}{$|{\cal D}_t| = 100$} & \multicolumn{3}{c}{$|{\cal D}_t| = 1000$} \\ \cmidrule(r){3-5} \cmidrule(r){6-8}
 &      & \ET\,(et)     & \FI\,(fi)     & \HU\,(hu)      & \ET\,(et)      & \FI\,(fi)     & \HU\,(hu)      \\ \midrule
       \multirow{3}{*}{\rotatebox{90}{source}}
 &   \ET\,(et)          & ---    & {\bf 60.9\%} & {\bf 60.4\%}  & ---     & {\bf 85.1\%} & 74.8\%  \\
 &   \FI\,(fi)          & {\bf 60.1\%} & ---    & 60.3\%  & {\bf 82.3\%}  & ---    & {\bf 75.2\%}  \\
 &   \HU\,(hu)          & 47.1\% & 48.3\% & ---     & 76.9\%  & 81.2\% & ---  \\ \midrule
& multi-source & 54.7\% & 55.3\% & 55.4\%  & 78.7\%  & 81.8\% & 73.3\% \\ \bottomrule
    \end{tabular}
    \end{adjustbox}
    \caption{Results for the Uralic languages.}
    \end{subtable}
    \caption{Results under our joint model. 
    The tables highlight that
    the best source languages are often genetically and typologically closest. Also, we see that multi-source often helps, albeit more often in the $|{\cal D}_t|=100$ case.    \label{tab:results}\looseness=-1}
\end{table*}

\subsubsection{Higher-Order CRF Tagger}
The {\sc MarMoT} tagger is
the leading non-neural approach to morphological tagging. This baseline is important since non-neural, feature-based approaches have
been found empirically to be more efficient, in the sense that their
learning curves tend to be steeper.
Thus, in the low-resource setting
we would be remiss not to consider a feature-based approach. 
Note that this is not a transfer approach, but rather only uses
the low-resource data.

\subsubsection{Alignment-based Projection}
The projection approach of \newcite{buys-botha:2016:P16-1} provides an
alternative method for transfer learning. The idea is to construct
pseudo-annotations for bitext given an alignment
\cite{och2003systematic}.  Then, one trains a standard tagger using
the projected annotations. The specific tagger employed is the {\sc wsabie}
model of \newcite{DBLP:conf/ijcai/WestonBU11}, which---like our
approach---is a $0^\text{th}$-order discriminative neural model. In
contrast to ours, however, their network is shallow. We compare
the two methods in more detail in \cref{sec:related-work}.

\begin{table*}
  \begin{adjustbox}{width=2.\columnwidth}
    \begin{tabular}{lccccccccccccc} \toprule
    &  \multicolumn{13}{c}{Accuracy} \\
      & \multicolumn{3}{c}{B\&B (2016)} & \multicolumn{2}{c}{{\sc MarMoT}} & \multicolumn{2}{c}{Ours (Mono)} & \multicolumn{2}{c}{Ours (Universal)}  & \multicolumn{2}{c}{Ours (Joint)}  & \multicolumn{2}{c}{Ours (Specific)} \\ \cmidrule(r){2-4} \cmidrule(r){5-6} \cmidrule(r){7-8} \cmidrule(r){9-10} \cmidrule(r){11-12} \cmidrule(r){13-14}  
              & en (int)  & best (non) & best (int) & $100$ & $1000$ & $100$ & $1000$ & $100$ & $1000$ & $100$ & $1000$ & $100$ & $1000$  \\ \cmidrule(r){2-14}
    \BG\,(bg) & 36.3      & 38.2       & 50.0       & 56.5 & 78.8 & 40.2 & 66.6 & 57.8 & 80.9   & 64.5 & {\bf 81.6}  & 63.5   & 80.8 \\
    \CS\,(cs) & 24.4      & 49.3       & 53.4       & 49.2 & 69.2 & 32.1 & 66.1 & 57.4 & 77.6   & 57.9 & {\bf 74.8}  & 56.1   & 74.2 \\
    \DA\,(da) & 36.6      & 46.9       & 46.9       & 75.9 & 90.9 & 45.3 & 86.6 & 77.6 & 90.1   & 87.8 & 93.9        & 89.2   & {\bf 94.3} \\
    \ES\,(es) & 39.9      & 75.3       & 75.5       & 85.9 & 93.1 & 64.7 & 92.5 & 85.1 & 60.9   & 90.9 & {\bf 94.2}  & 90.7   & {\bf 94.2} \\
    \FI\,(fi) & 27.4      & 51.8       & 56.0       & 50.0 & 77.5 & 28.0 & 74.2 & 48.3 & 81.2   & 55.3 & 81.8        & 55.4   & 80.7 \\
    \IT\,(it) & 38.1      & 75.5       & 75.9       & 81.7 & 92.3 & 67.0 & 88.9 & 84.7 & 93.1   & 86.8 & {\bf 93.8}  & 86.1   & 93.3 \\
    \PL\,(pl) & 25.3      & 47.4       & 51.3       & 51.7 & 71.1 & 32.1 & 60.9 & 47.4 & {\bf 78.4}   & 57.0 & 78.1        & 56.1   & 76.4 \\
    \PT\,(pt) & 36.6      & 71.9       & 72.2       & 77.0 & 86.3 & 61.7 & 85.6 & 80.6 & 88.7   & 83.4 & 88.7        & 82.4   & {\bf 89.1} \\
    \SV\,(sv) & 29.3      & 44.5       & 44.5       & 69.5 & 88.3 & 46.1 & 84.2 & 75.7 & 90.0   & 77.2 & {\bf 90.9}  & 78.3   & 90.7 \\
 \bottomrule
  \end{tabular}
  \end{adjustbox}
  \caption{Comparison of our approach to various baselines for low-resource tagging under token-level accuracy. We compare on only those languages in \newcite{buys-botha:2016:P16-1}. Note that tag-level accuracy was not reported in the original B\&B paper, but was acquired through personal communication with the first author. All architectures presented in this work are used in their multi-source setting. The B\&B and {\sc MarMoT} models are single-source. }
  \label{tab:baseline-table1}
\end{table*}

\subsubsection{Architecture Study}
Additionally, we perform a thorough study
of the neural transfer learner, considering
all three architectures. A primary goal of our experiments
is to determine which of our three proposed neural transfer techniques
is superior. Even though our experiments focus on morphological tagging,
these architectures are more general in that they may be easily applied to
other tasks, e.g., parsing or machine translation. 
We additionally explore the viability of multi-source transfer, i.e.,
the case where we have multiple source languages. All of our architectures
generalize to the multi-source case without any complications.

\subsection{Experimental Details}\label{sec:details}
We train our models with the following conditions. 
\paragraph{Evaluation Metrics.}
 We evaluate using average per-token accuracy, as is standard for
 both POS tagging and morphological tagging, and per-feature $F_1$ as
 employed in \newcite{buys-botha:2016:P16-1}. The per-feature $F_1$
calculates a key $F^k_1$ for each key in the target language's tags
 by asking if the key--attribute pair $k_j{=}v_j$ is in the predicted
 tag.
 Then, the key-specific $F^k_1$ values are averaged equally.
 Note that $F_1$ is a more flexible metric as it gives partial
 credit for getting some of the attributes in the bundle correct,
 whereas accuracy does not.

 \paragraph{Hyperparameters.}
 Our networks are four layers deep (two LSTM layers for the character embedder, i.e., to compute $\mathbf{v}_n$ and two LSTM layers for the tagger,
 i.e., to compute $\mathbf{e}_n$) 
 and we use a representation size of 128 for
 the character input vector size and hidden layers of 256 nodes in all other cases. All networks are trained with the
 stochastic gradient method RMSProp \cite{Tieleman2012}, with a fixed
 initial learning rate and a learning rate decay that is adjusted for
 the other languages according to the amount of training data. The
 batch size is always 16. Furthermore, we use dropout
 \cite{DBLP:journals/jmlr/SrivastavaHKSS14}. The dropout probability
 is set to 0.2. We used Torch 7 \cite{collobert2011torch7} to configure the computation graphs
 implementing the network architectures.

\section{Results and Discussion}
We report our results in two tables. First, we report a detailed
cross-lingual evaluation in \cref{tab:results}.  Secondly, we report a
comparison against two baselines in \cref{tab:baseline-table1} (accuracy)
and \cref{tab:baseline-table2} ($F_1$). We
see two general trends of the data. First, we find that genetically closer
languages yield better source languages. Second, we find that the
multi-softmax architecture is the best in terms of transfer ability,
as evinced by the results in \cref{tab:results}. We find a wider gap between our model
and the baselines under the accuracy than under
$F_1$. We attribute this to the fact
that $F_1$ is a softer metric in that
it assigns credit to partially correct guesses.

\paragraph{Source Language.}
As discussed in \cref{sec:transferring-morphology}, the transfer of
morphology is language-dependent. This intuition is borne out in the results from our
study (see \cref{tab:results}).  We see that in the closer grouping of the Western Romance
languages, i.e., Catalan, French, Italian, Portuguese, and Spanish, it
is easier to transfer than with Romanian, an Eastern Romance language.
Within the Western grouping, we see that the close
pairs, e.g., Spanish and Portuguese, are amenable to
transfer. We find a similar pattern in the other language families,
e.g., Russian is the best source language for Ukrainian, Czech is the best
source language for Slovak and Finnish is the best source language for Estonian.

\paragraph{Multi-Source Transfer.}
 In many cases, we
find that multiple sources noticeably improve the results over the
single-source case. For instance, when we have multiple Romance languages
as a source language, we see gains of up to 2\%. We also see
gains in the Northern Germanic languages when
using multiple source languages. From a linguistic point of view,
this is logical as different source languages may be similar to the
target language along different dimensions, e.g., when transferring among the Slavic languages, we note that Russian retains the complex
nominal case system of Ukrainian, but south Slavic Bulgarian is lexically more similar.

\paragraph{Performance Against the Two Baselines.}
As shown in \cref{tab:baseline-table1} and \cref{tab:baseline-table2}, our model outperforms the projection tagger of
\newcite{buys-botha:2016:P16-1} even though our approach does not
utilize bitext, large-scale alignment or
monolingual corpora---rather, all transfer
between languages happens through the forced sharing of character-level
features.\footnote{
We would like to highlight some issues of comparability
with the results in \newcite{buys-botha:2016:P16-1}.
Strictly speaking, the results are not comparable
and our improvement over their method should be
taken with a grain of salt. As the source code
is not publicly available and developed in industry, we resorted to numbers
in their published work and additional numbers obtained
through direct communication with the authors.
First, we used a slightly newer version of UD to incorporate
more languages: we used v2 whereas they used v1.2. There
are minor differences in the morphological tagset used between these
versions. Also, in the $|{\cal D}_t|=1000$ setting, we are training
on significantly more data than the models in \newcite{buys-botha:2016:P16-1}.
A much fairer comparison is to our models with $|{\cal D}_t| = 100$.
Also, we compare to their method using their standard (non) setup.
This method is fair in so far as we evaluate in the same manner, but it disadvantages
their approach, which cannot predict tags that are not in the source language.\looseness=-1}
Our model does, however, require annotation of a small
number of sentences in the target language for training. We note,
however, that this does not necessitate a large number of human annotation hours \cite{garrette-baldridge:2013:NAACL-HLT}.

\begin{table*}
  \begin{adjustbox}{width=2.\columnwidth}
    \begin{tabular}{lccccccccccccc} \toprule
      &  \multicolumn{13}{c}{$F_1$} \\
      & \multicolumn{3}{c}{B\&B (2016)} & \multicolumn{2}{c}{{\sc MarMoT}} & \multicolumn{2}{c}{Ours (Mono)} & \multicolumn{2}{c}{Ours (Universal)}  & \multicolumn{2}{c}{Ours (Joint)}  & \multicolumn{2}{c}{Ours (Specific)} \\ \cmidrule(r){2-4} \cmidrule(r){5-6} \cmidrule(r){7-8} \cmidrule(r){9-10} \cmidrule(r){11-12} \cmidrule(r){13-14}
              & en (int)  & best (non) & best (int) & $100$ & $1000$ & $100$   & $1000$  & $100$ & $1000$ & $100$ & $1000$       & $100$  & $1000$  \\ \cmidrule(r){2-14}
    \BG\,(bg) & 51.6      & 61.9       & 65.0       & 53.7  & 74.7   & 26.0    & 68.0    & 55.1  & 77.3   & 56.6  & 77.8         & 55.1   & {\bf 78.6} \\
    \CS\,(cs) & 55.7      & 61.6       & 64.0       & 60.8  & 80.5   & 30.9    & 65.3    & 54.5  & 66.3   & 54.7  & 66.5         & 54.6   & {\bf 67.0} \\
    \DA\,(da) & 65.4      & 70.7       & 73.1       & 69.7  & 92.9   & 35.3    & 90.1    & 85.9  & 93.2   & 86.9  & {\bf 93.5}         & 83.2   & 93.2 \\
    \ES\,(es) & 60.7      & 74.0       & 74.6       & 82.4  & 92.6   & 55.9    & 91.4    & 88.4  & 93.6   & 89.2  & {\bf 94.1}         & 87.6   & 93.8 \\
    \FI\,(fi) & 59.1      & 57.2       & 59.1       & 44.6  & 78.3   & 17.5    & 61.7    & 48.6  & 73.6   & 49.3  & {\bf 74.4}         & 46.2   & 73.9 \\
    \IT\,(it) & 66.1      & 74.4       & 75.3       & 78.7  & 90.0   & 56.4    & 87.0    & 83.1  & 90.5   & 83.3  & {\bf 91.9}         & 82.7   & 91.7 \\
    \PL\,(pl) & 47.3      & 56.8       & 60.4       & 57.8  & 81.8   & 31.6    & 69.7    & 61.9  & 83.9   & 62.5  & {\bf 84.7}         & 62.6   & 83.2 \\
    \PT\,(pt) & 60.2      & 69.2       & 73.1       & 67.6  & 80.0   & 42.9    & 82.0    & 77.9  & 86.3   & 78.1  & {\bf 86.5}         & 71.8   & 85.7 \\
    \SV\,(sv) & 55.1      & 72.1       & 74.6       & 69.7  & 90.2   & 44.1    & 86.4    & 82.5  & 93.2   & 83.5  & {\bf 93.7}         & 82.8   & 93.4 \\
 \bottomrule
  \end{tabular}
  \end{adjustbox}
  \caption{Comparison of our approach to various baselines for low-resource tagging under $F_1$ to allow for a more complete
    comparison to the model of \newcite{buys-botha:2016:P16-1}. All architectures presented in this work are used in their multi-source setting. The B\&B and {\sc MarMoT} models are single-source. We only compare on those languages used in B\&B.}
  \label{tab:baseline-table2}
\end{table*}

\begin{figure}
  \centering
  \includegraphics[width=7.5cm]{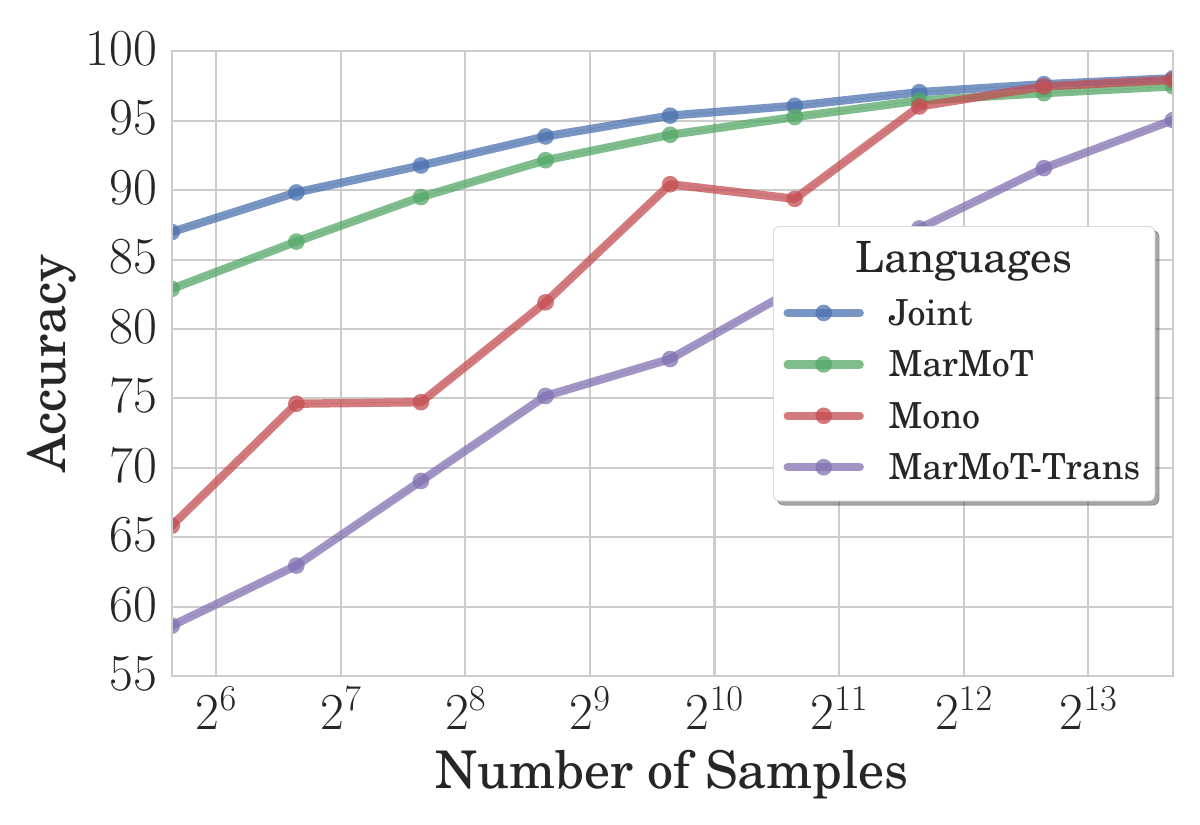}
    \caption{A learning curve for Spanish and Catalan comparing
      our monolingual model, our joint model, and two {\sc MarMoT} models. 
      The
      first {\sc MarMoT} model is identical to those trained in the rest of the paper
      and the second shows a multi-task approach, which failed so no further experimentation was performed with this model.\looseness=-1}
      \label{fig:curve}
\end{figure}

\paragraph{Reducing Sample Complexity.}
Another interesting point
about our model that is best evinced in \cref{fig:curve}
is the feature-based CRF approach seems to be
a better choice for the low-resource setting, i.e.,
the neural model has greater sample complexity. However,
in the multi-task scenario, we find that
the neural tagger's learning curve is even steeper.
In other words, if we have to train a tagger on very little
data, we are better off using a neural multi-task approach
than a feature-based approach; preliminary
attempts to develop a multi-task version of {\sc MarMoT}
failed (see \cref{fig:curve}).



\section{Related Work}\label{sec:related-work}
We divide the discussion of related work topically into three parts
for ease of intellectual digestion.

\subsection{Alignment-Based Distant Supervision}
Most cross-lingual work in NLP---focusing on morphology or
otherwise---has concentrated on indirect supervision, rather than transfer
learning. The goal in such a scenario is to provide noisy labels
for training the tagger in the low-resource language through
annotations projected over aligned bitext with a high-resource
language. This method of projection was first introduced by
\newcite{DBLP:conf/naacl/YarowskyN01} for the projection of POS
annotation. While follow-up work
\cite{DBLP:conf/ijcnlp/FossumA05,das-petrov:2011:ACL-HLT2011,tackstrom-mcdonald-uszkoreit:2012:NAACL-HLT}
has continually demonstrated the efficacy of projecting simple part-of-speech annotations,
\newcite{buys-botha:2016:P16-1} were the first to show the use
of bitext-based projection for the training of a {\em morphological}
tagger for low-resource languages. 

As we also discuss the training of a morphological tagger, our work is
most closely related to \newcite{buys-botha:2016:P16-1} in terms of
the task itself. We contrast the approaches. The main difference lies
therein, that our approach is not projection-based and, thus, does not
require the construction of a bilingual lexicon for projection-based
on bitext.  Rather, our method jointly learns multiple taggers and
forces them to share features---a true transfer learning scenario. In
contrast to projection-based methods, our procedure always requires a
minimal amount of annotated data in the low-resource target
language---in practice, however, this distinction is non-critical as
projection-based methods without a small amount of seed target language
data perform poorly \cite{buys-botha:2016:P16-1}.

\subsection{Character-level NLP}
Our work also follows a recent trend in NLP, whereby traditional
word-level neural representations are being replaced by
character-level representations for a myriad of tasks, e.g., POS tagging
\cite{DBLP:conf/icml/SantosZ14}, parsing
\cite{ballesteros-dyer-smith:2015:EMNLP}, language modeling
\cite{ling-EtAl:2015:EMNLP2}, sentiment analysis
\cite{DBLP:conf/nips/ZhangZL15} as well as the tagger of
\newcite{heigold2017}, whose work we build upon.  Our work is also
related to recent work on character-level morphological generation using neural architectures
\cite{faruqui-EtAl:2016:N16-1,rastogi-cotterell-eisner:2016:N16-1}.\looseness=-1

\subsection{Neural Cross-lingual Transfer in NLP}
In terms of methodology, however, our proposal bears similarity
to recent work in speech and machine translation---we discuss
each in turn. In speech recognition, \newcite{heigold2013multilingual} train
a cross-lingual neural acoustic model on five Romance languages.
The architecture bears similarity to our multi-language softmax
approach. Dependency parsing benefits from cross-lingual learning in a similar fashion \cite{guo-EtAl:2015:ACL-IJCNLP2,DBLP:conf/aaai/GuoCYWL16}.

In neural machine translation
\cite{DBLP:conf/nips/SutskeverVL14,DBLP:journals/corr/BahdanauCB14},
recent work
\cite{firat-cho-bengio:2016:N16-1,zoph-knight:2016:N16-1,JohnsonSLKWCTVW16}
has explored the possibility of jointly train translation models for a
wide variety of languages. Our work addresses a different task, but
the undergirding philosophical motivation is similar, i.e.,
attack low-resource NLP through multi-task transfer learning.
\newcite{kann-cotterell-schutze:2017:ACL2017} offer a similar method for cross-lingual
transfer in morphological inflection generation.

\section{Conclusion}
We have presented three character-level recurrent neural network architectures for
multi-task cross-lingual transfer of morphological
taggers.  We provided an empirical evaluation of the technique on 18
languages from four different language families, showing wide-spread
applicability of the method. We found that the transfer of morphological
taggers is an eminently viable endeavor among related languages and,
in general, the closer the languages, the easier the transfer of
morphology becomes. Our technique outperforms two strong baselines
proposed in previous work. Moreover, we define standard low-resource
training splits in UD for future research in low-resource morphological tagging.
Future work should focus on extending the neural morphological tagger
to a joint lemmatizer \cite{muller-EtAl:2015:EMNLP} and evaluate its functionality in the low-resource setting.

\section*{Acknowledgements}
RC acknowledges the support of an NDSEG fellowship. Also, we
would like to thank Jan Buys and Jan Botha who helped us compare to the numbers reported in their paper. We would also like to thank Hinrich Sch{\"u}tze, Tim Vieira and Jason Naradowsky for
reading early drafts.

\bibliography{crosslingual-neural-tagging}
\bibliographystyle{emnlp_natbib}

\end{document}